\documentclass[runningheads]{llncs}
\usepackage{epstopdf}
 
\usepackage[]{eccv}

\usepackage{eccvabbrv}

\usepackage{graphicx}
\usepackage{booktabs}
\usepackage{xcolor}
\usepackage[normalem]{ulem}
\useunder{\uline}{\ul}{}

\usepackage[accsupp]{axessibility}  

\usepackage[pagebackref,breaklinks,colorlinks,citecolor=eccvblue]{hyperref}

\usepackage{orcidlink}
\usepackage{graphicx}
\usepackage{booktabs}
\usepackage{listings}
\usepackage{multicol}
\usepackage{multirow}
\usepackage{fancyhdr}
\titlerunning{ }
\authorrunning{ }

\begin{document}

\title{MSRAMIE: Multimodal Structured Reasoning Agent for Multi-instruction Image Editing} 

\author{Zhaoyuan Qiu\inst{1}\orcidlink{0009-0004-3054-5287} \and
Ken Chen\inst{1}\orcidlink{0009-0009-9247-7722} \and
Xiangwei Wang\inst{1}\orcidlink{0000-0003-3069-9214} \and
Yu Xia \inst{1}\orcidlink{0000-0001-7014-2926} \and
Sachith Seneviratne \inst{1}\orcidlink{0000-0001-9094-2736} \and
Saman Halgamuge \inst{1}\orcidlink{0000-0002-2536-4930}
}

\institute{University Of Melbourne, Grattan Street, Parkville, Victoria 3010, Australia \\
\email{zhaoyuanqiu@student.unimelb.edu.au}}

\maketitle 

\begin{abstract}
Existing instruction-based image editing models perform well with simple, single-step instructions but degrade in realistic scenarios that involve multiple, lengthy, and interdependent directives. A main cause is the scarcity of training data with complex multi-instruction annotations. However, it is costly to collect such data and retrain these models. To address this challenge, we propose \textbf{MSRAMIE}, a training-free agent framework built on Multimodal Large Language Model (MLL\\M). MSRAMIE takes existing editing models as plug-in components and handle multi-instruction tasks via \textbf{structured multimodal reasoning}. It orchestrates iterative interactions between an MLLM-based \textit{Instructor} and an image editing \textit{Actor}, introducing a novel reasoning topology that comprises the proposed \textbf{Tree-of-States} and \textbf{Graph-of-References}. During inference, complex instructions are decomposed into multiple editing steps which enable state transitions, cross-step information aggregation, and original input recall, which enables systematic exploration of the image editing space and flexible progressive output refinement. The visualizable inference topology further provides interpretable and controllable decision pathways. Experiments show that as the instruction complexity increases, MSRAMIE can improve instruction following over 15\% and increases the probability of finishing all modifications in a single run over 100\%, while preserving perceptual quality and maintaining visual consistency. 

\end{abstract}

\begin{figure}[tb]
  \centering
  \includegraphics[height=3.8cm]{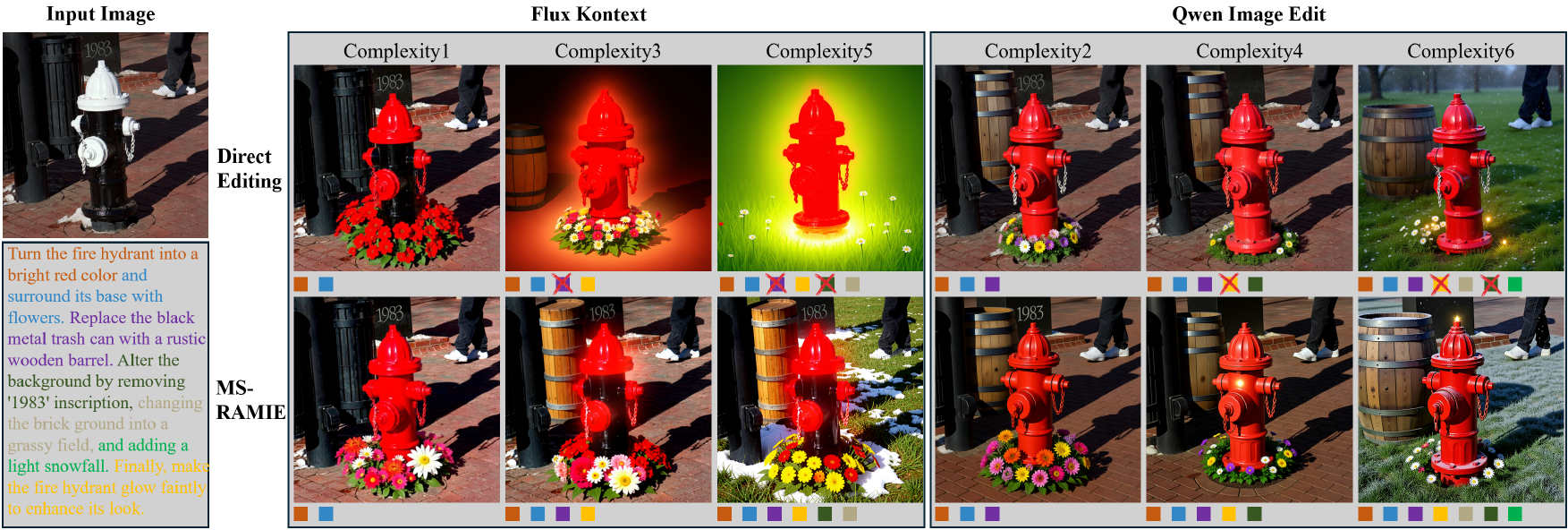}
  \caption{Performance Degradation Under Multi-instruction. Existing models degrade when directly dealing with lengthy multi-instructions, manifested by poor instruction following and identity preservation that worsen with increasing complexity (1st row). MSRAMIE mitigates these issues (2nd row). Colored text under the input image stands for different parts of the editing multi-instructions, corresponding to color blocks under each image. A red cross stands for a failed instruction.
  }
  \label{fig:Fig1}
\end{figure}

\section{Introduction}
\label{sec:intro}
\textbf{Instruction-based image editing}, introduced by InstructPix2Pix~\cite{brooks2023instructpix2pix}, established a user-friendly paradigm by fine-tuning models such as Stable Diffusion 1.5~\cite{rombach2022latent} to follow natural language instructions, offering greater convenience than previous approaches~\cite{zhang2023controlnet,hu2022lora,ruiz2023dreambooth}. Subsequent work~\cite{zhang2023magicbrush,zhao2024ultraedit,ge2024seeddataedit,bai2024humanedit,jiang2025anyedit} released high-quality datasets to further advance the field. Based on these data and methodological foundations, recent large-scale models~\cite{wu2025qwenimage,flux2techreport,bfl2025fluxkontext,wu2025chronoedit,liu2025step1xedit,deng2025bagel,team2025longcat} achieve state-of-the-art performance, demonstrating strong ability to generate high-fidelity and semantically coherent results under single instructions~\cite{jia2025compbench}. However, real-world editing tasks often require multiple modifications at the same time, resulting in \textbf{long and complex multi-instruction} inputs. Complex-Edit~\cite{yang2025complexedit} reports a consistent performance degradation as instruction complexity increases. This limitation reflects a fundamental gap: open-source datasets~\cite{brooks2023instructpix2pix,zhang2023magicbrush,zhao2024ultraedit,ge2024seeddataedit,bai2024humanedit,jiang2025anyedit} predominantly contain short or single-step instructions, leading to poor generalization under multi-instruction scenarios, as shown in \Cref{fig:Fig1}. While collecting multi-instruction datasets and retraining large-scale models are costly, a training-free solution is needed.

Practical solutions often rely on \textbf{repeated resampling} or \textbf{iterative refinement} of intermediate results. Nevertheless, these strategies are limited: (1) they are labor-intensive and impractical for large-scale deployment; (2) repeated text interaction imposes a burden on users, as effective prompting often requires prompt-engineering expertise~\cite{sahoo2024promptsurvey}; and (3) most editing models are designed for single-round input-output mapping and lack mechanisms to maintain history, retrieve prior decisions, or preserve user intent across iterations~\cite{ma2025talk2image}, making them unsuitable for multi-round editing.

To address these challenges, we propose \textbf{MSRAMIE}, a fully automated, training-free agent framework for multi-instruction image editing. It integrates existing editing models as plug-in components and handles multi-instructions through a novel multimodal structured reasoning paradigm, which solves the mentioned problems of existing models during multi-round editing. Specifically, we introduce a hybrid reasoning topology comprising a \textbf{Tree-of-States (ToS)} and a \textbf{Graph-of-References (GoR)}. Such a paradigm decomposes the task into a series of structured problem-solving steps, which allows automatic prompt generation, flexible state transitions, information aggregation and input recall. Experimental results show that, in multi-instruction scenarios, MSRAMIE significantly improves instruction following and identity preservation while maintaining perceptual quality of the generation results. Our contributions include:

\begin{enumerate}
    \item We introduce MSRAMIE, a fully automated and training-free multimodal agent framework that augments existing instruction-based image editing models to handle multi-instruction inputs without additional retraining.
    \item We propose the structured multimodal reasoning paradigms Tree-of-States (ToS) and Graph-of-References (GoR) for instruction-based image editing, which greatly facilitates the solution search in the image editing space. It also enables a transparent, trackable, and controllable editing process through the visualizable inference process.
\end{enumerate}

\begin{figure}[ht]
  \centering
  \includegraphics[height=4cm]{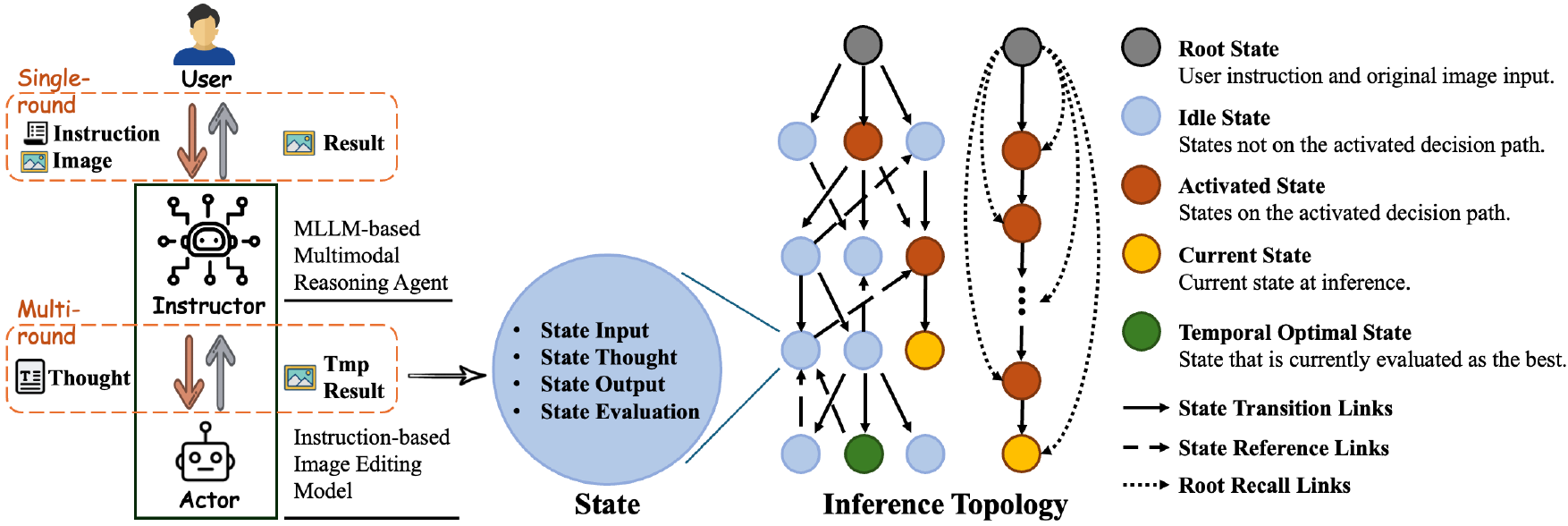}
  \caption{MSRAMIE Architecture. Iterative interactions between the \textit{Instructor} and \textit{Actor} progressively construct an inference topology, where each state corresponds to one round of interaction. Transition and reference links among states form a \textbf{Tree-of-States} and a \textbf{Graph-of-References}, enabling multiple decision paths, or \textbf{Chain-of-States}, containing progressively refined candidate solutions in the image editing space. All states retain access to the root state for the original inputs during their creation.
  }
  \label{fig:Fig2}
\end{figure}

\begin{figure}[h]
  \centering
  \includegraphics[height=6cm]{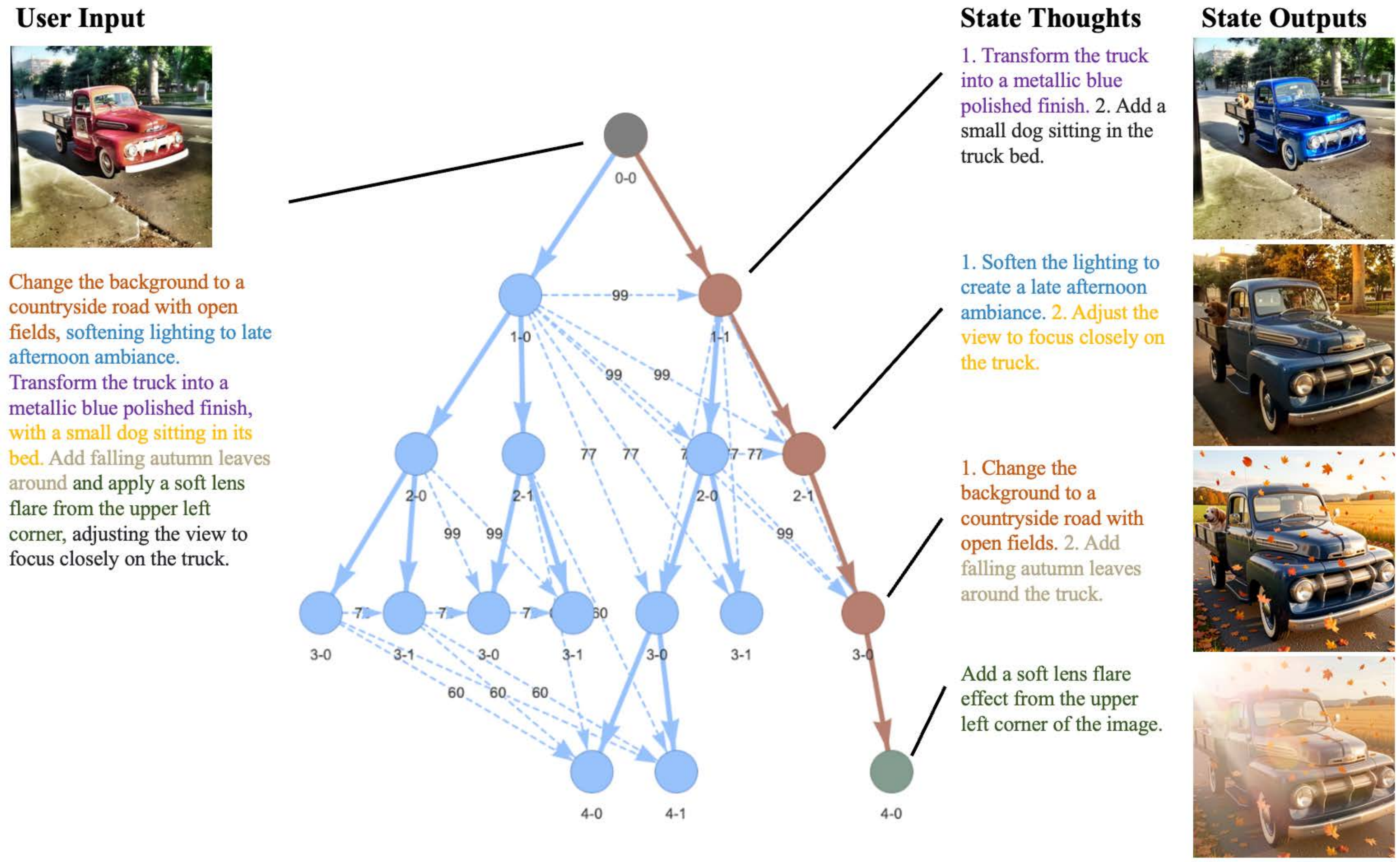}
  \caption{Image Editing Space Exploration by Structured Multimodal Reasoning. MSRAMIE creates a visualizable reasoning topology. States on a decision path dynamically focus on different parts of the original lengthy multi-instruction, with referencing to nearby states to prevent duplicated exploration. Inference topology expands in depth-first style, providing a transparent and controllable editing process.
  }
  \label{fig:Fig3}
\end{figure}

\section{Related Works}
We provide a basic review of related methods and attempts to integrate them with instruction-based image generation and editing.

\subsection{MLLMs in Instruction-based Image Editing}
Multimodal Large Language Models (MLLMs) integrate visual perception with high-level textual reasoning ability. Due to this, MLLMs~\cite{bai2023qwenvl,wang2024qwen2vl,bai2025qwen25vl,bai2025qwen3vl,liu2023llava,shahriar2024gpt4o,geminiteam2023gemini,geminiteam2024gemini15} have been widely applied to complex vision-language tasks, including instruction-based image editing~\cite{fu2023mgie,huang2024smartedit,fang2025got,liu2025step1xedit}. CompBench~\cite{jia2025compbench} highlights MLLMs as critical for high-performance editing, emphasizing their role in bridging textual instructions and visual understanding through multimodal reasoning. However, most existing approaches depend on curated training or fine-tuning, while training-free integration strategies remain largely underexplored.

\subsection{Structured Multimodal Reasoning in Instruction-based Image Editing}
Structured prompting proves to enhance LLMs in complex tasks by imposing explicit reasoning topologies that guide inference in a controllable manner~\cite{besta2024topology}. Representative paradigms, such as Chain-of-Thought (CoT)~\cite{wei2022cot}, Tree-of-Thought (ToT)~\cite{yao2023tot}, and Graph-of-Thought (GoT)~\cite{besta2024got}, introduce progressively richer reasoning structures and substantially improve performance in complex tasks.

These ideas have been extended to multimodal settings, leading to structured multimodal reasoning frameworks, which are conceptually similar but greatly differ in the form of presentation. Multimodal Chain-of-Thought (MCoT)\cite{wang2025mcot}, the simplest form, is widely used in instruction-based image editing~\cite{yang2025complexedit,fang2025got,zhuo2023reflection,guo2025cotimg}. While these studies demonstrate the benefits of chain-structured reasoning for instruction-based image editing, more expressive tree-structured and graph-structured reasoning schemas remains unexplored.

\subsection{MLLM-based Image Generation/Editing Agent}
With the rapid advancement of MLLMs, MLLM-based agent systems have been increasingly adopted for complex tasks to enable automatic planning and problem solving. For instruction-based image generation and editing, recent studies explored the integration of chain-structured reasoning\cite{huang2025dialoggen,wang2024genartist,ye2025genpilot,ma2025talk2image,ye2026agentbanana}. They naturally exhibit stepwise execution agent behaviors, showing inherent compatibility with structured reasoning. However, the integration of tree-structured or graph-structured multimodal reasoning frameworks remains unexplored.

\section{Method}
\subsection{Task Definition}
\subsubsection{Evaluation Dimensions \& Metrics}
\label{sec:Metrics}
We follow Complex-Edit\cite{yang2025complexedit} in the evaluation protocol of instruction-based image editing tasks. The evaluation system consists of three dimensions:
\begin{enumerate}
    \item \textbf{Instruction Following (IF)}: Measures how well the edited images implement the requested modification in the user instruction. For this dimension, we use \textbf{VQAScore}\cite{lin2024vqascore}, which formulates the evaluation as a visual question answering (VQA) task and computes the proportion of affirmative answers. In addition, we report \textbf{Complete Instruction Following (CIF)}, defined as the fraction of samples with VQAScore equal to one, indicating compliance with complete editing requirements.
    \item \textbf{Identity Preservation (IP)}: Evaluates whether visual elements of the input image that are expected to remain unchanged are successfully preserved. For this dimension, we adopt \textbf{CLIP-I}\cite{radford2021clip}, which is robust to geometric transformations such as scaling or rotation.
    \item \textbf{Perceptual Quality (PQ)}: Assesses overall visual quality, including perceptual realism, texture fidelity, diversity, \etc. For this dimension, we calculate the \textbf{Fréchet Inception Distance (FID)}\cite{heusel2017fid} between the original and edited image sets.
\end{enumerate}

It is important to note that instruction-based image editing inherently involves a trade-off between IF and IP: as the extent of editing increases, the edited image may deviate further from the original input. Consequently, improvements in one dimension may come at the expense of the other.

Moreover, PQ is often positively correlated with IP, particularly when editing non-synthesized images. Producing visually realistic content with fine-grained details remains a major challenge for current models\cite{falck2025fourierdiffusion,gruszczynski2025beyondblur}. These models are also prone to loss of detail and structural distortion due to encoding images to latent representations and subsequent decoding\cite{berrada2025perceptualldm}. As a result, a higher degree of IP typically leads to a less degraded PQ.

\subsubsection{Task Description}
Given an initial image and the editing multi-instruction which contains many sub-instructions from the user and a limited inference budget, the goal is to search for an optimal image within the image editing space while maintaining specific constraints on IF, IP and PQ. Through structured multimodal reasoning, MSRAMIE transforms the original task into searching a trajectory $\tau^*$ that leads to an optimal solution in the constraint editing space:
\begin{equation}
\tau^{*} = \arg\max_{\tau \in T_{B}(I_{0}, x)} \left[ \lambda_{1} \sum_{i=1}^{n} C(I_{T}, x_{i}) + \lambda_{2} P(I_{T}, I_{0}) + \lambda_{3} Q(I_{T}) \right],
\end{equation}
\begin{equation*}
\text{s.t.} \quad I_{t+1} = f(I_{t}, a_{t})
\end{equation*}
, where $I_0$ denotes the initial input image and $I_t$ denotes the intermediate result of the t-th step on a trajectory $\tau$ with length $T$.
$x_i$ denotes a single instruction contained in the user-provided editing multi-instruction. $a_t$ denotes the editing performed in $I_t$ and $f$ denotes the editing model.
$C$, $P$, $Q$, respectively, denotes the evaluation methods of IF, IP, and PQ. 
$T_B$ denotes the trajectories set in the editing space constraint by the inference budget $B$.

\subsection{Framework Design}
MSRAMIE comprises two top-level components: an MLLM-based \textit{\textbf{Instructor}} module and an \textit{\textbf{Actor}} module. The typical usage scenario is illustrated in \cref{fig:Fig2}, in which the system interacts with a human user in a single round, while internally conducting multi-round interactions between the two modules to derive the final editing result. During inference, the underlying structured reasoning process progressively constructs an inference topology that integrates a Tree-of-States with a Graph-of-References. This provides a transparent editing process exemplified in \cref{fig:Fig3}. We explain the terminologies in the following sections.

\subsubsection{State: Abstraction of One Interaction Round}
Each interaction round is represented as a state. The \textit{Instructor} retrieves relevant past states and generates a textual ``thought”. The \textit{Actor} receives the intermediate image from the previous round and edits it based on ``thought”. The edited result is then returned to the \textit{Instructor} for evaluation. Each state is characterized by four attributes:
\begin{itemize}
    \item State input: the editing result of the previous round.
    \item State output: the editing result of the current round, as a candidate solution.
    \item State thought: the \textit{Instructor}-generated textual prompt to guide the \textit{Actor}.
    \item State evaluation: evaluation reasoning and quantitative metric scores produced by the \textit{Instructor}.
\end{itemize}

\subsubsection{Inference Topology: Abstraction of Structured Reasoning-based Multi-round Interaction}
The multi-round interaction process is formalized as the progressive expansion of an inference topology. Each state in this topology corresponds to a round of \textit{Instructor}-\textit{Actor} interaction except the root state, which represents the initial user interaction. The root state takes the original image for both its input and output, and the original multi-instruction as the state thought. Intuitively, root state has no evaluation.

At each inference step, a new state is created and appended to the inference topology. The process ends when a completion state is reached, the inference budget is exhausted, or the topology expansion limitation is reached. The optimal state output will be returned based on customized scheduling rules as the final editing result. This design decomposes complex editing requests into a series of sub-tasks, with each state yielding a candidate solution and the path from the root defining a decision trajectory. Depending on whether it is on the final selected trajectory, a state is either ``activated'' or ``idle''.

To enhance exploration, the inference topology has a dual-component design: a Tree-of-States and a Graph-of-References. These two structured multimodal reasoning schemas increase the likelihood and efficiency of finding a globally optimal solution by incorporating resampling, traceback, and aggregation mechanisms into the reasoning process.

\subsubsection{Tree-of-States (ToS)}
\label{sec:ToS}
Because a single decision path may converge to a locally optimal solution due to the high variance inherent in stochastic sampling, a resampling and traceback mechanism is required to identify superior alternatives. We formalize the passing of state inputs and outputs as \textbf{State Transition Links} in the inference topology, where each state receives input from exactly one predecessor and may transition to multiple successors. This structure induces an N-ary tree topology that governs the state evolution process during inference, termed Tree-of-States (ToS).

In addition, each state logically maintains direct access to the original user instruction and original image throughout its creation procedure. By doing so, it guarantees a consistent reference for instruction following and identity preservation. We formalize this access to the root state as \textbf{Root Recall Links}.

Finally, the output of any state represents an aggregation of the preceding states along the path from the root to the current node. These states form an activated decision trajectory, or \textbf{Chain-of-States (CoS)}, for the task and provide explicit, stepwise evidence of how the agent arrives at its current solution.

\subsubsection{Graph-of-References (GoR)}
\label{sec:GoR}
In the early development stage, we observed that states sharing the same parent often generated highly similar thoughts. This behavior produced nearly duplicated states along different decision paths and resulted in a waste of the inference budget. To improve search efficiency in the image editing space, it is necessary to have a mechanism allowing reference to the nearby similar states before thought generating.

Information retrieval among states is formalized as \textbf{State Reference Links} in the inference topology. Each state can access previous states in a predefined search range, thereby forming a graph structure that enables local aggregation of thoughts from neighboring states, termed Graph-of-References (GoR). 

After retrieving the reasoning outputs of nearby states with similar inputs, the current state's thought generating is encouraged to focus on non-covered aspects of these reference states. Ablation tests demonstrate that this strategy improves the quality of the final solution under the same limited inference budget, as reported in \Cref{table:Table2} and \Cref{table:Table3}.
\subsection{Modules Design}

\begin{figure}[tb]
  \centering
  \includegraphics[height=5.3cm]{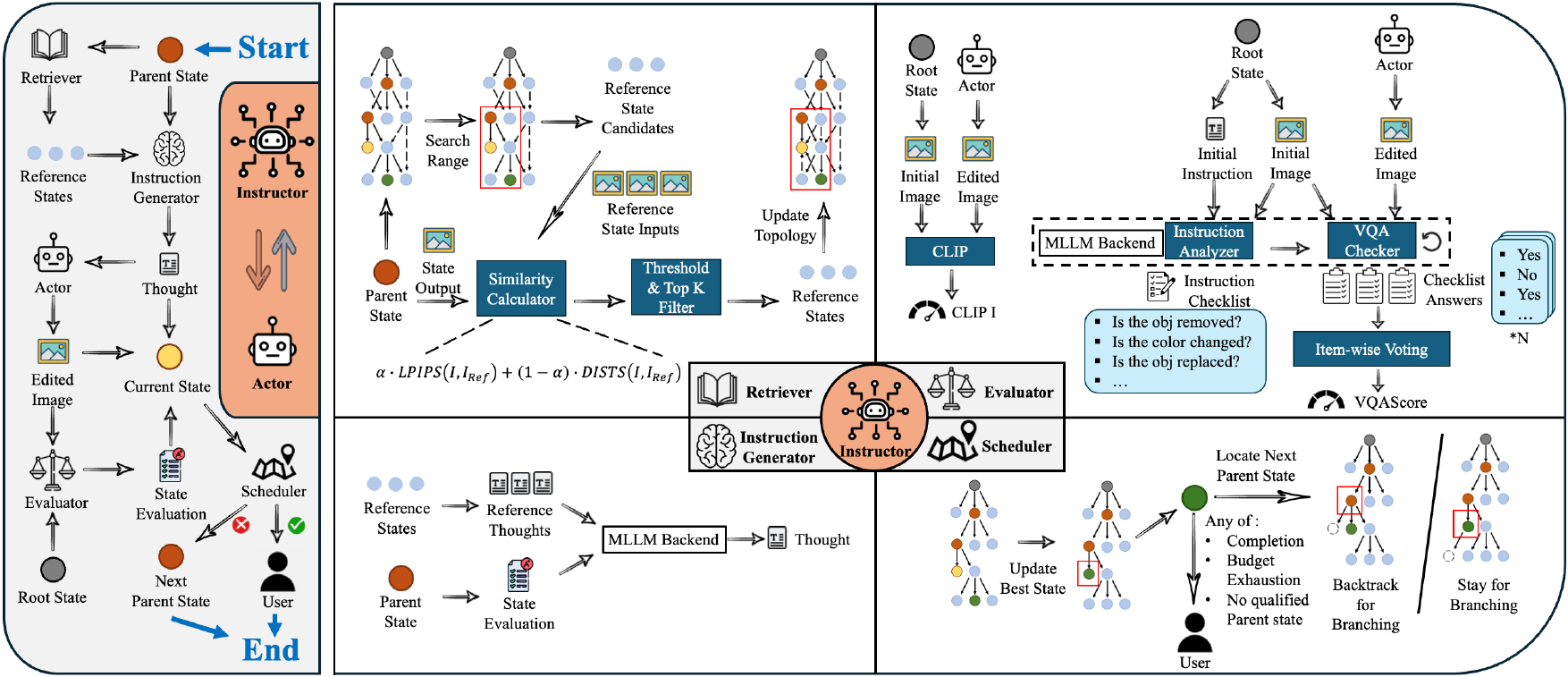}
  \caption{\textit{Instructor}-\textit{Actor} Interaction Process (left) and \textit{Instructor} Submodules Process (right). Four \textit{Instructor} submodules coordinate with the \textit{Actor} to create a new state. One may refer to \Cref{fig:Fig2} for  the meaning of state colors.
  }
  \label{fig:Fig4}
\end{figure}

\subsubsection{\textit{Instructor} \& \textit{Actor}}
The \textit{Actor} module is a plug-in instruction-based image editing model and can be freely replaced by any compatible implementation.

The \textit{Instructor} module is a structured reasoning system comprising four submodules: (1) \textbf{Retriever}: identifies similar states in the inference topology and extracts their associated thoughts as references for the current state. (2) \textbf{Instruction Generator}: produces the thought for the current state using a customizable MLLM backend. (3) \textbf{Evaluator}: assesses the output of the current state using predefined metrics, also implemented with a customizable MLLM backend. (4) \textbf{Scheduler}: manages the exploration of the inference topology, determines whether an acceptable candidate solution has been found, and decides whether to terminate inference or continue searching.

For each newly created state, these submodules coordinate with the \textit{Actor} to generate a candidate solution as demonstrated in \Cref{fig:Fig4}: (1) the parent state is passed to the Retriever to obtain reference states and their thoughts; (2) together with the parent state evaluation, reference states provide the Instruction Generator with their state thoughts to produce a new thought; (3) the \textit{Actor} takes in the thought and perform editing to the parent state output and generate an updated image; (4) the Evaluator scores this output with respect to the original user instruction; and (5) the Scheduler determines whether to terminate inference and return the best state, or to continue by selecting an existing state in the topology as the next parent state. The inner process of submodules will be elaborated in the following sections.

\subsubsection{Retriever}
The Retriever submodule, operates in four sequential steps: (1) it receives the parent state and retrieves its state output, which serves as the input to the current state; (2) then identifies all existing states within a predefined search range in the inference topology as candidate reference states; (3) it computes the similarity between each candidate and the parent state, selecting the top Kmost similar states while discarding those below a minimum similarity threshold; and (4) the selected reference states are forwarded to the Instruction Generator.

State similarity is defined by the image similarity between each candidate reference state input and the parent state’s output. The similarity score is computed as a weighted average of DISTS\cite{ding2020dists} and LPIPS\cite{zhang2018lpips}, which capture structural and perceptual visual similarity and thus favor reference states with closely matched visual composition.

\subsubsection{Evaluator}
The Evaluator submodule provides real-time assessment of each state along Identity Preservation (IP) and Instruction Following (IF), as illustrated in the top-right of \Cref{fig:Fig4}.

For IP evaluation, original and edited images are encoded with a pretrained CLIP model\cite{radford2021clip}, and CLIP-I is computed as the embeddings cosine similarity.

For IF evaluation, the Evaluator operates as follows: (1) it retrieves the original user instruction from the root state. The instruction is decomposed into a checklist of VQA questions using an Instruction Analyzer, where each question aims at a specific required visual modification; (2) a VQAChecker then answers each question with a binary response (“Yes”/“No”), repeating the query N times to improve robustness; (3) final answers are obtained by aggregating repeated responses, and VQAScore is calculated as the proportion of affirmative answers.

The Instruction Analyzer and VQAChecker share the same MLLM backend to reduce memory use. To ensure stable and structured generation, we employ Chain-of-Thought prompting \cite{wei2022cot} and guided decoding\cite{willard2023guideddecoding,zheng2024sglang}, with prompt templates provided in Appendix E. We initially explored CLIP-T as an IF metric; however, pretrained CLIP models\cite{radford2021clip} exhibit degraded reliability for lengthy text, motivating the use of VQAScore for more reliable evaluation.

\subsubsection{Instruction Generator}
The Instruction Generator, operates in two sequential steps: (1) it receives the parent state, retrieves its evaluation results, and extracts the associated VQA questions and answers, while incorporating thoughts from the retrieved reference states; and (2) it generates a new state thought that prioritizes unmet editing requirements identified in the parent state. Reference thoughts are included to encourage explorations of alternative reasoning paths.

To improve search efficiency and reduce perceptual degradation from repeated image encoding and decoding, we introduce an \textbf{Instruction Volume (IV)} parameter. At each iteration, the Instruction Generator produces a thought containing up to IV sub-instructions to guide the \textit{Actor}. Larger IV values enable higher instruction satisfaction at shallower depths in the inference topology. 

The Instruction Generator is implemented with an MLLM backend using guided decoding\cite{willard2023guideddecoding,zheng2024sglang} to enforce structured output formatting. Prompt templates are provided in Appendix E.

\subsubsection{Scheduler}
When the current state is passed to the Scheduler, the following procedure is executed, which is illustrated in the bottom-right of \Cref{fig:Fig4}: (1)
Topology update. The current state is added to the inference topology, and the temporally optimal state is updated according to a configurable ranking rule, which in our implementation prioritizes VQAScore followed by CLIP-I; (2) Budget check. If the number of states in the topology has reached the computational limit, inference terminates and the temporally optimal state is returned; otherwise, the process continues; (3) Completion check. The temporally optimal state is evaluated against the completion criteria, including metric thresholds and a minimum depth requirement. If satisfied, inference terminates; otherwise, the process proceeds; (4) Stay-condition evaluation. The current state is tested against four constraints: minimum metric thresholds, depth below the maximum limit, bounded degradation relative to the parent state, and metric scores below the completion upper bound. If satisfied, the state becomes the parent for the next state; and (5) Backtracking. If the current state fails the stay conditions, the Scheduler backtracks to a shallower parent state with remaining expansion capacity; if none exists up to the root, inference terminates and the temporally optimal state is returned. This induces a depth-first expansion of the topology.

When multiple temporally optimal states share the same evaluation score, the system selects the state at the shallowest depth that exceeds the minimum depth threshold, thereby reducing PQ degradation from repeated encoding and decoding. If multiple optimal states occur at the same depth, all are returned to the user for selection.

\section{Experiments}
\subsection{Dataset}
\begin{figure}[hb]
  \centering
  \includegraphics[height=3.8cm]{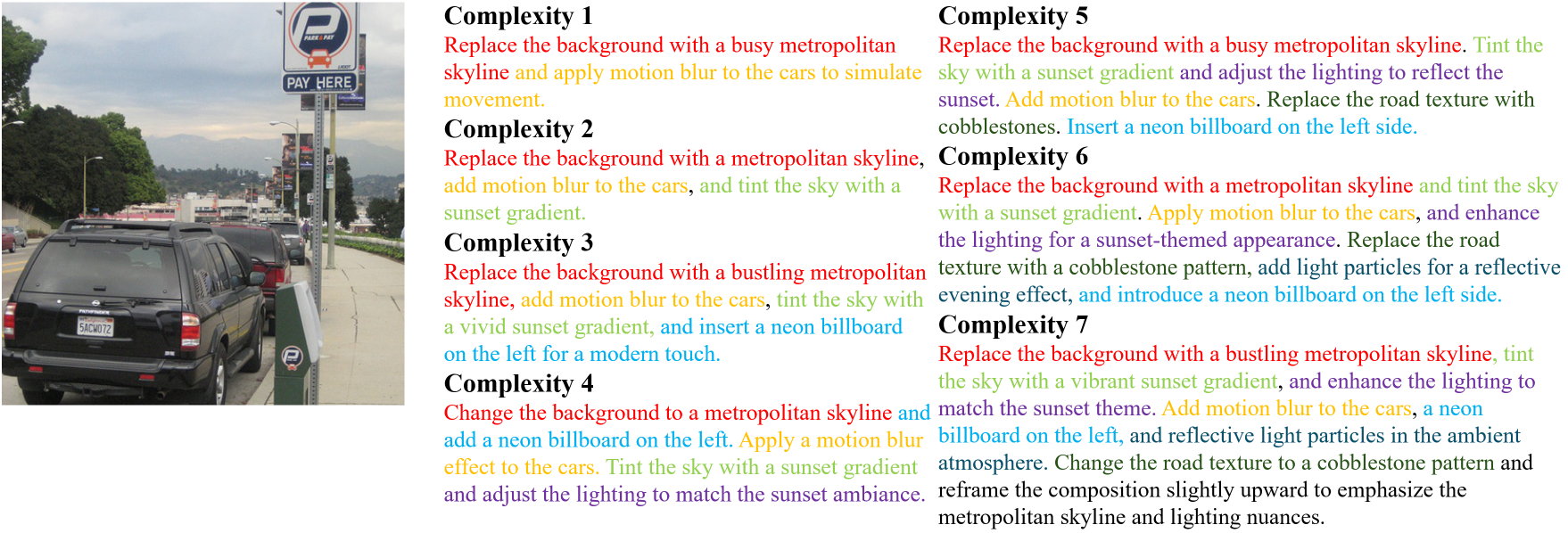}
  \caption{A Sample of Complex-Edit. Each non-synthetic image is accompanied with lengthy multi-instructions of 7 complexity levels. Higher complexity leads to more and lengthier editing requirements.
  }
  \label{fig:Fig5}
\end{figure}

We use the real image dataset of Complex-Edit\cite{yang2025complexedit}, an open-source dataset for the multi-instruction image editing task. Its subsets C1-7 contains 531 non-synthetic images paired with textual multi-instructions of different complexity levels. A sample is shown in \Cref{fig:Fig5}.

\subsection{Experiment Setup}
For setup, we select the open-source Qwen3-VL\cite{bai2025qwen3vl} as the MLLM basis of the \textit{Instructor} module. For the \textit{Actor} module, we run experiments with the pioneering open-source instruction-based image editing models, including Qwen-image-edit\cite{wu2025qwenimage}, Flux-kontext\cite{bfl2025fluxkontext}, Flux2-Klein\cite{flux2techreport}. The hyperparameter settings are in Appendix D. All experiments were carried out on an H100 GPU. 

\subsection{Quantitative Results}
Following the three evaluation dimensions introduced in \Cref{sec:Metrics}, IF, IP and PQ, we compare direct editing with existing models against MSRAMIE-augmented editing, quantifying the improvements with selected metrics.

As shown in \Cref{table:Table1}, for direct editing, we observe that: the performance in all three evaluation dimensions degrades as the complexity of the instruction increases, suggesting a general degradation of existing models under multi-instruction scenarios. For MSRAMIE-augmented editing, we observe that: (1) substantial improvements are achieved in IF, which becomes more pronounced at higher levels of complexity; (2) modest improvements are achieved in IP, which are relatively consistent between complexity levels; and (3) minor differences are made in PQ, indicating that the generation quality of basic models is largely preserved. Moreover, an increasing trend of FID with higher complexity is observed in FluxKlein, which is due to the accumulated loss of detail during repeated image encoding and decoding with deeper inference. This will be discussed in Appendix B. The complete test results are provided in Appendix AMai.

Overall, our method consistently improves existing models in multi-instruction scenarios, with substantial gains in instruction following while maintaining visual consistency and generation quality, suggesting the great potential of structured multimodal reasoning for image editing tasks.
\begin{table}[!h]
  \caption{Performance enhancement by MSRAMIE. Baseline models exhibit consistent performance degradation as instruction complexity increases. In contrast, MSRAMIE delivers progressively greater IF (VQAScore \& CIF) improvements under higher instruction complexity, with enhanced IP (CLIP-I) and largely preserved PQ (FID). This demonstrates its effectiveness in challenging multi-instruction scenarios.
}
  \label{table:Table1}
  \fontsize{5.5pt}{5.5pt}\selectfont

\begin{tabular}{@{}lccccccccc@{}}
\toprule
\multicolumn{10}{c}{VQAScore (↑)} \\ \midrule
\multicolumn{1}{|l|}{Models}     
& \multicolumn{3}{c|}{QwenImageEdit}                                            
& \multicolumn{3}{c|}{FluxKontext}                                              
& \multicolumn{3}{c|}{FluxKlein} \\ \midrule

\multicolumn{1}{|l|}{Complexity} 
& Direct & MSRAMIE & \multicolumn{1}{c|}{Improvement} 
& Direct & MSRAMIE & \multicolumn{1}{c|}{Improvement} 
& Direct & MSRAMIE & \multicolumn{1}{c|}{Improvement} \\ \cmidrule(r){1-1}

\multicolumn{1}{|l|}{1}          
& 0.8088 & {\ul \textbf{0.8760}} & \multicolumn{1}{c|}{8.31\%}    
& 0.7884 & {\ul \textbf{0.8659}} & \multicolumn{1}{c|}{9.84\%}    
& 0.8898 & {\ul \textbf{0.9315}} & \multicolumn{1}{c|}{4.68\%}    \\

\multicolumn{1}{|l|}{2}          
& 0.8041 & {\ul \textbf{0.8947}} & \multicolumn{1}{c|}{11.27\%}   
& 0.7779 & {\ul \textbf{0.8692}} & \multicolumn{1}{c|}{11.74\%}   
& 0.9134 & {\ul \textbf{0.9441}} & \multicolumn{1}{c|}{3.36\%}    \\

\multicolumn{1}{|l|}{3}          
& 0.7795 & {\ul \textbf{0.8854}} & \multicolumn{1}{c|}{13.59\%}   
& 0.7667 & {\ul \textbf{0.8705}} & \multicolumn{1}{c|}{13.54\%}   
& 0.9134 & {\ul \textbf{0.9543}} & \multicolumn{1}{c|}{4.49\%}    \\

\multicolumn{1}{|l|}{4}          
& 0.7530 & {\ul \textbf{0.8768}} & \multicolumn{1}{c|}{16.44\%}   
& 0.7331 & {\ul \textbf{0.8599}} & \multicolumn{1}{c|}{17.30\%}   
& 0.8934 & {\ul \textbf{0.9383}} & \multicolumn{1}{c|}{5.03\%}    \\

\multicolumn{1}{|l|}{5}          
& 0.7412 & {\ul \textbf{0.8870}} & \multicolumn{1}{c|}{19.67\%}   
& 0.7445 & {\ul \textbf{0.8651}} & \multicolumn{1}{c|}{16.19\%}   
& 0.8953 & {\ul \textbf{0.9491}} & \multicolumn{1}{c|}{6.00\%}    \\

\multicolumn{1}{|l|}{6}          
& 0.7325 & {\ul \textbf{0.8756}} & \multicolumn{1}{c|}{19.53\%}   
& 0.7326 & {\ul \textbf{0.8568}} & \multicolumn{1}{c|}{16.95\%}   
& 0.8888 & {\ul \textbf{0.9406}} & \multicolumn{1}{c|}{5.82\%}    \\

\multicolumn{1}{|l|}{7}          
& 0.7172 & {\ul \textbf{0.8780}} & \multicolumn{1}{c|}{22.41\%}   
& 0.7361 & {\ul \textbf{0.8599}} & \multicolumn{1}{c|}{16.82\%}   
& 0.8904 & {\ul \textbf{0.9482}} & \multicolumn{1}{c|}{6.48\%}    \\ \midrule

\multicolumn{1}{|l|}{Overall}    
& 0.7623 & {\ul \textbf{0.8819}} & \multicolumn{1}{c|}{15.69\%}   
& 0.7542 & {\ul \textbf{0.8639}} & \multicolumn{1}{c|}{14.55\%}   
& 0.8978 & {\ul \textbf{0.9437}} & \multicolumn{1}{c|}{5.12\%}    \\ \midrule
\addlinespace
\addlinespace
\toprule
\multicolumn{10}{c}{CIF(↑)} \\ \midrule
\multicolumn{1}{|l|}{Models}     
& \multicolumn{3}{c|}{QwenImageEdit}                                            
& \multicolumn{3}{c|}{FluxKontext}                                              
& \multicolumn{3}{c|}{FluxKlein} \\ \midrule

\multicolumn{1}{|l|}{Complexity} 
& Direct & MSRAMIE & \multicolumn{1}{c|}{Improvement} 
& Direct & MSRAMIE & \multicolumn{1}{c|}{Improvement} 
& Direct & MSRAMIE & \multicolumn{1}{c|}{Improvement} \\ \cmidrule(r){1-1}

\multicolumn{1}{|l|}{1}          
& 0.6064 & {\ul \textbf{0.7307}} & \multicolumn{1}{c|}{20.50\%}   
& 0.5838 & {\ul \textbf{0.7213}} & \multicolumn{1}{c|}{23.55\%}   
& 0.7476 & {\ul \textbf{0.8362}} & \multicolumn{1}{c|}{11.84\%}   \\

\multicolumn{1}{|l|}{2}          
& 0.5292 & {\ul \textbf{0.7306}} & \multicolumn{1}{c|}{38.08\%}   
& 0.4783 & {\ul \textbf{0.6667}} & \multicolumn{1}{c|}{39.37\%}   
& 0.7571 & {\ul \textbf{0.8362}} & \multicolumn{1}{c|}{10.45\%}   \\

\multicolumn{1}{|l|}{3}          
& 0.4011 & {\ul \textbf{0.6478}} & \multicolumn{1}{c|}{61.50\%}   
& 0.3917 & {\ul \textbf{0.6121}} & \multicolumn{1}{c|}{56.25\%}   
& 0.7006 & {\ul \textbf{0.8267}} & \multicolumn{1}{c|}{18.01\%}   \\

\multicolumn{1}{|l|}{4}          
& 0.3032 & {\ul \textbf{0.6008}} & \multicolumn{1}{c|}{98.14\%}   
& 0.3107 & {\ul \textbf{0.5235}} & \multicolumn{1}{c|}{68.48\%}   
& 0.6045 & {\ul \textbf{0.7495}} & \multicolumn{1}{c|}{23.99\%}   \\

\multicolumn{1}{|l|}{5}          
& 0.2241 & {\ul \textbf{0.5405}} & \multicolumn{1}{c|}{141.18\%}  
& 0.2542 & {\ul \textbf{0.4859}} & \multicolumn{1}{c|}{91.11\%}   
& 0.5650 & {\ul \textbf{0.7665}} & \multicolumn{1}{c|}{35.67\%}   \\

\multicolumn{1}{|l|}{6}          
& 0.2166 & {\ul \textbf{0.4840}} & \multicolumn{1}{c|}{123.48\%}  
& 0.2015 & {\ul \textbf{0.4407}} & \multicolumn{1}{c|}{118.69\%}  
& 0.4878 & {\ul \textbf{0.7081}} & \multicolumn{1}{c|}{45.17\%}   \\

\multicolumn{1}{|l|}{7}          
& 0.1676 & {\ul \textbf{0.4708}} & \multicolumn{1}{c|}{180.90\%}  
& 0.1714 & {\ul \textbf{0.4068}} & \multicolumn{1}{c|}{137.36\%}  
& 0.4482 & {\ul \textbf{0.6780}} & \multicolumn{1}{c|}{51.26\%}   \\ \midrule

\multicolumn{1}{|l|}{Overall}    
& 0.3497 & {\ul \textbf{0.6008}} & \multicolumn{1}{c|}{71.77\%}   
& 0.3417 & {\ul \textbf{0.5510}} & \multicolumn{1}{c|}{61.26\%}   
& 0.6158 & {\ul \textbf{0.7716}} & \multicolumn{1}{c|}{22.29\%}   \\ \midrule
\addlinespace
\addlinespace
\toprule
\multicolumn{10}{c}{CLIP-I(↑)} \\ \midrule
\multicolumn{1}{|l|}{Models}     
& \multicolumn{3}{c|}{QwenImageEdit}                                            
& \multicolumn{3}{c|}{FluxKontext}                                              
& \multicolumn{3}{c|}{FluxKlein} \\ \midrule

\multicolumn{1}{|l|}{Complexity} 
& Direct & MSRAMIE & \multicolumn{1}{c|}{Improvement} 
& Direct & MSRAMIE & \multicolumn{1}{c|}{Improvement} 
& Direct & MSRAMIE & \multicolumn{1}{c|}{Improvement} \\ \cmidrule(r){1-1}

\multicolumn{1}{|l|}{1}          
& 0.9532 & {\ul \textbf{0.9575}} & \multicolumn{1}{c|}{0.45\%}    
& 0.9457 & {\ul \textbf{0.9514}} & \multicolumn{1}{c|}{0.61\%}    
& 0.9442 & {\ul \textbf{0.9495}} & \multicolumn{1}{c|}{0.56\%}    \\

\multicolumn{1}{|l|}{2}          
& 0.9410 & {\ul \textbf{0.9446}} & \multicolumn{1}{c|}{0.38\%}    
& 0.9325 & {\ul \textbf{0.9381}} & \multicolumn{1}{c|}{0.60\%}    
& 0.9305 & {\ul \textbf{0.9363}} & \multicolumn{1}{c|}{0.62\%}    \\

\multicolumn{1}{|l|}{3}          
& 0.9356 & {\ul \textbf{0.9380}} & \multicolumn{1}{c|}{0.26\%}    
& 0.9248 & {\ul \textbf{0.9286}} & \multicolumn{1}{c|}{0.41\%}    
& 0.9214 & {\ul \textbf{0.9277}} & \multicolumn{1}{c|}{0.68\%}    \\

\multicolumn{1}{|l|}{4}          
& 0.9289 & {\ul \textbf{0.9322}} & \multicolumn{1}{c|}{0.36\%}    
& 0.9149 & {\ul \textbf{0.9199}} & \multicolumn{1}{c|}{0.55\%}    
& 0.9151 & {\ul \textbf{0.9202}} & \multicolumn{1}{c|}{0.56\%}    \\

\multicolumn{1}{|l|}{5}          
& 0.9221 & {\ul \textbf{0.9249}} & \multicolumn{1}{c|}{0.31\%}    
& 0.9121 & {\ul \textbf{0.9167}} & \multicolumn{1}{c|}{0.50\%}    
& 0.9095 & {\ul \textbf{0.9136}} & \multicolumn{1}{c|}{0.45\%}    \\

\multicolumn{1}{|l|}{6}          
& 0.9146 & {\ul \textbf{0.9181}} & \multicolumn{1}{c|}{0.38\%}    
& 0.9072 & {\ul \textbf{0.9095}} & \multicolumn{1}{c|}{0.26\%}    
& 0.9060 & {\ul \textbf{0.9079}} & \multicolumn{1}{c|}{0.21\%}    \\

\multicolumn{1}{|l|}{7}          
& 0.9106 & {\ul \textbf{0.9190}} & \multicolumn{1}{c|}{0.91\%}    
& 0.9071 & {\ul \textbf{0.9114}} & \multicolumn{1}{c|}{0.47\%}    
& {\ul \textbf{0.9024}} & {0.9015} & \multicolumn{1}{c|}{-0.10\%}   \\ \midrule

\multicolumn{1}{|l|}{Overall}    
& 0.9294 & {\ul \textbf{0.9335}} & \multicolumn{1}{c|}{0.43\%}    
& 0.9206 & {\ul \textbf{0.9251}} & \multicolumn{1}{c|}{0.49\%}    
& 0.9185 & {\ul \textbf{0.9224}} & \multicolumn{1}{c|}{0.43\%}    \\ \midrule
\addlinespace
\addlinespace

\toprule
\multicolumn{10}{c}{FID(↓)} \\ \midrule
\multicolumn{1}{|l|}{Models}     
& \multicolumn{3}{c|}{QwenImageEdit}                                            
& \multicolumn{3}{c|}{FluxKontext}                                              
& \multicolumn{3}{c|}{FluxKlein} \\ \midrule

\multicolumn{1}{|l|}{Complexity} 
& Direct & MSRAMIE & \multicolumn{1}{c|}{Improvement} 
& Direct & MSRAMIE & \multicolumn{1}{c|}{Improvement} 
& Direct & MSRAMIE & \multicolumn{1}{c|}{Improvement} \\ \cmidrule(r){1-1}

\multicolumn{1}{|l|}{1}          
& 75.61 & {\ul \textbf{72.48}}  & \multicolumn{1}{c|}{4.14\%}   
& 84.37 & {\ul \textbf{80.90}}  & \multicolumn{1}{c|}{4.11\%}   
& 84.14 & {\ul \textbf{79.88}}  & \multicolumn{1}{c|}{5.06\%}   \\

\multicolumn{1}{|l|}{2}          
& 88.38 & {\ul \textbf{88.18}}  & \multicolumn{1}{c|}{0.23\%}   
& 103.40 & {\ul \textbf{102.41}} & \multicolumn{1}{c|}{0.96\%}   
& 96.16 & {\ul \textbf{95.88}}  & \multicolumn{1}{c|}{0.29\%}   \\

\multicolumn{1}{|l|}{3}          
& {\ul \textbf{98.49}} & {99.73}  & \multicolumn{1}{c|}{-1.26\%}    
& {\ul \textbf{116.21}} & {118.94} & \multicolumn{1}{c|}{-2.35\%}    
& 105.59 & {\ul \textbf{103.23}} & \multicolumn{1}{c|}{2.24\%}   \\

\multicolumn{1}{|l|}{4}          
& 121.70 & {\ul \textbf{121.85}} & \multicolumn{1}{c|}{-0.12\%}    
& 136.98 & {\ul \textbf{136.20}} & \multicolumn{1}{c|}{0.57\%}   
& {\ul \textbf{119.09}} & {121.78} & \multicolumn{1}{c|}{-2.26\%}    \\

\multicolumn{1}{|l|}{5}          
& 141.18 & {\ul \textbf{139.87}} & \multicolumn{1}{c|}{0.93\%}   
& {\ul \textbf{142.71}} & {142.92} & \multicolumn{1}{c|}{-0.15\%}    
& {\ul \textbf{123.63}} & {127.66} & \multicolumn{1}{c|}{-3.26\%}    \\

\multicolumn{1}{|l|}{6}          
& {\ul \textbf{140.13}} & {146.20} & \multicolumn{1}{c|}{-4.33\%}    
& {\ul \textbf{156.12}} & {157.65} & \multicolumn{1}{c|}{-0.98\%}    
& {\ul \textbf{133.14}} & {135.29} & \multicolumn{1}{c|}{-1.61\%}    \\

\multicolumn{1}{|l|}{7}          
& 150.05 & {\ul \textbf{148.35}} & \multicolumn{1}{c|}{1.13\%}   
& 167.44 & {\ul \textbf{167.04}} & \multicolumn{1}{c|}{0.24\%}   
& {\ul \textbf{144.15}} & {153.12} & \multicolumn{1}{c|}{-6.22\%}    \\ \midrule

\multicolumn{1}{|l|}{Overall}    
& {\ul \textbf{116.51}} & 116.69 & \multicolumn{1}{c|}{-0.14\%}   
& 129.60 & {\ul \textbf{129.43}} & \multicolumn{1}{c|}{0.13\%}    
& {\ul \textbf{115.13}} & {116.66} & \multicolumn{1}{c|}{-1.36\%}    \\ \midrule
\end{tabular}
\\
\end{table}

\subsection{Qualitative Results}
\begin{figure}[!h]
  \centering
  \includegraphics[height=9cm]{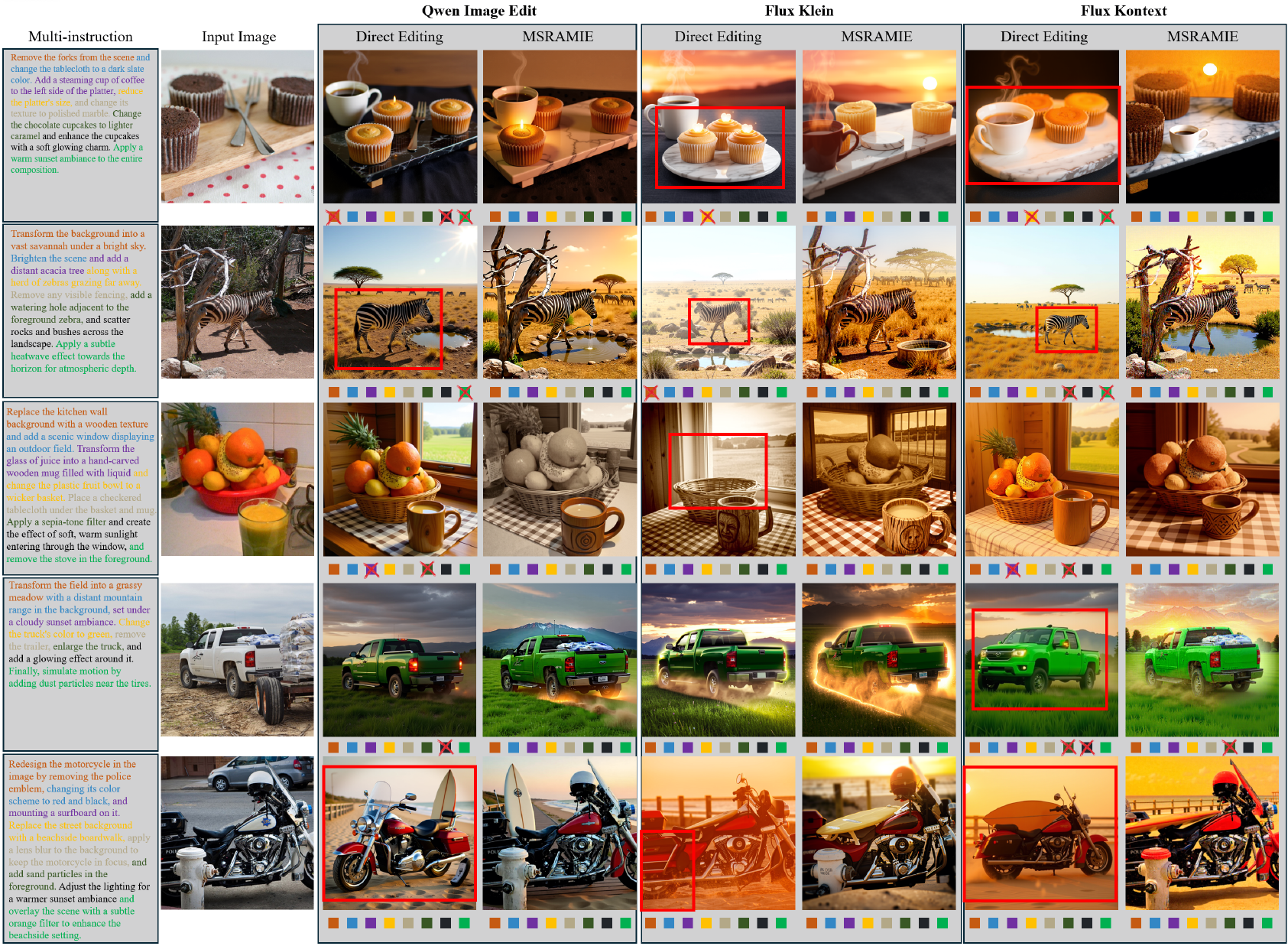}
  \caption{Direct Editing vs. MSRAMIE-augmented Editing. Colored blocks beneath each image denote individual instruction items, and a red cross indicates failure to satisfy the corresponding item. Regions where visual content of original images is not preserved in Direct Editing but retained by MSRAMIE are highlighted with red boxes.
  }
  \label{fig:Fig6}
\end{figure}

We present visual comparisons between direct editing outputs and MSRAMIE-generated results to illustrate the improvements as shown in \Cref{fig:Fig6}.

For IF, existing models often miss parts of the lengthy multi-instructions, whereas MSRAMIE captures them and improves the compliance. For IP, existing models may substantially repaint the image, failing to preserve visual elements not explicitly specified (composition, geometric structure, object details, \etc). MSRAMIE markedly improves IP by retaining them. For PQ, MSRAMIE shows no noticeable degradation in perceptual quality compared to direct editing.

These improvements can be attributed to the divide-and-conquer strategy in the structured reasoning process: task decomposition allows models that perform well on short instructions to operate more effectively. Shorter sub-instructions enable the model to focus on regions requiring modification, thereby protecting unaffected content from unwanted change. The ToS and GoR topologies further enhance performance by reducing inference variance and improving search efficiency. The preserved perceptual quality comes as a by-product of improved IP due to their positive correlation.

\subsection{Inference-scaling Study}
We study the effects of inference budget and depth by varying the number of inference steps and evaluating states at different depths. The results show that increasing both the budget and the depth consistently improves the final solution but leads to a trade-off among IF, IP and PQ. Details are in Appendix B.
\subsection{Ablation Study}
We isolate and evaluate the contributions of key mechanisms by comparing direct editing, Resampling-Only editing, Chain-of-States-Only (CoS-Only) editing, Tree-of-States-Only (ToS-Only) editing, and the full MSRAMIE framework.

As shown in \Cref{table:Table2}, Resampling-Only and CoS-Only yield modest gains in IF, whereas ToS-Only provides substantial improvement by allowing flexible state transitions within structured reasoning. MSRAMIE further enhances IF by incorporating GoR. Reasons are intuitive: Resampling-Only reduces sampling variance but is limited by shallow inference; CoS-Only allows for deeper inference but suffers from high variance in single sampling; ToS-Only combines their strengths while mitigating the weaknesses; and GoR further improves performance by avoiding redundant exploration and improving search efficiency.

As shown in \Cref{table:Table3}, for IP, CoS-Only has little effect, whereas Resampling-Only, ToS-Only, and full MSRAMIE yield progressively larger gains. This is achieved by the gradually broadened editing space exploration. For PQ, CoS-Only and ToS-Only degrade the quality, while Resampling-Only improves it; combining them in MSRAMIE mitigates these degradations and results in a modest overall gain. Intuitively, the high sampling variance of CoS-Only accumulates loss and distortion of detail, whereas Resampling-Only improves visual consistency and quality via a best-of-N strategy. Incorporating GoR in MSRAMIE further offsets the drawbacks of CoS-Only by improving exploration efficiency.

Involved methods explanation and results details are in Appendix C.

\begin{table}[h]
  \caption{Instruction Following Ablation Test. Results aggregated across all instruction complexities and all samples, with improvements reported relative to direct editing.
  }
  \label{table:Table2}
  \centering
  \fontsize{5.5pt}{5.5pt}\selectfont
\begin{tabular}{@{}lccccc@{}}
\toprule
\textbf{Metrics} & \multicolumn{1}{l}{\textbf{Direct}} & \multicolumn{1}{l}{\textbf{Resampling-Only}} & \multicolumn{1}{l}{\textbf{CoS-Only}} & \multicolumn{1}{l}{\textbf{ToS-Only}} & \multicolumn{1}{l}{\textbf{MSRAMIE}} \\ \midrule
VQAScore(↑)         & 0.7623                              & 0.7917                                       & 0.8115                                & 0.8666                                & {\ul \textbf{0.8819}}                \\
Improvement      & -                                   & 3.85\%                                       & 6.50\%                                & 13.83\%                               & {\ul \textbf{15.89\%}}               \\ \midrule
CIF(↑)              & 0.3497                             & 0.4036                                       & 0.5056                               & 0.5895                                & {\ul \textbf{0.6008}}                \\
Improvement      & -                                   & 17.45\%                                      & 57.52\%                               & 90.19\%                               & {\ul \textbf{94.82\%}}               \\ \bottomrule
\end{tabular}
\end{table}

\begin{table}[h]
  \caption{Identity Preservation \& Perceptual Quality Ablation Test. Because of the trade-off among IF, IP, and PQ, IF is controlled in the IP/PQ ablation by selecting only samples from direct editing and MSRAMIE with identical VQAScore. Enhancement is then quantified by the CLIP-I percentage increase and the FID percentage decrease.
  }
  \label{table:Table3}
  \centering
  \fontsize{5.5pt}{5.5pt}\selectfont
\begin{tabular}{@{}l|cc|cc|cc|cc@{}}
\toprule
\textbf{Metrics} & \multicolumn{1}{l}{\textbf{Direct}} & \multicolumn{1}{l|}{\textbf{Resampling-Only}} & \multicolumn{1}{l}{\textbf{Direct}} & \multicolumn{1}{l|}{\textbf{CoS-Only}} & \multicolumn{1}{l}{\textbf{Direct}} & \multicolumn{1}{l|}{\textbf{ToS-Only}} & \multicolumn{1}{l}{\textbf{Direct}} & \multicolumn{1}{l}{\textbf{MSRAMIE}} \\ \midrule
CLIP-I(↑)           & 0.9370                              & 0.9388                                        & 0.9355                     & 0.9354                                 & 0.9333                     & 0.9365                                 & 0.9338                     & 0.9376                                \\
Improvement      & -                                   & 0.20\%                                        & -                          & 0.00\%                                 & -                          & 0.35\%                                 & -                          & {\ul \textbf{0.41\%}}                 \\ \midrule
FID(↓)              & 91.27                               & 90.89                                         & 107.54                     & 110.38                                 & 108.24                     & 108.81                                 & 107.91                     & 107.78                                \\
Improvement         & -                                   & {\ul\textbf{0.42\%}}                                       & -                          & -2.64\%                                 & -                          & -0.52\%                                & -                          & {\ul\textbf{0.12\%}}                \\ \bottomrule
\end{tabular}
\end{table}

\section{Conclusion}
We propose \textbf{MSRAMIE} for multi-instruction image editing tasks. It is a training-free agent framework with existing editing models as plug-in components. Based on the novel curated \textbf{Tree-of-States} and \textbf{Graph-of-References} structured multimodal reasoning schema, MSRAMIE improves basic models substantially on Instruction Following (IF), modestly on Identity Preservation (IP), and largely preserves their Perceptual Quality (PQ). Our work shows the great potential of  multimodal structured reasoning in complex image editing tasks.

\par\vfill\par


%
%
\bibliographystyle{splncs04}
\bibliography{main}

\clearpage 

\section*{Appendices}
\thispagestyle{empty}
\begin{tabular}{ll}
Appendix A & \hyperref[sec:AppendixA]{More Quantitative Analysis} \\
Appendix B & \hyperref[sec:AppendixB]{Inference-scaling Study} \\
Appendix C & \hyperref[sec:AppendixC]{Ablation Study Results Details} \\
Appendix D & \hyperref[sec:AppendixD]{Experiment Setup} \\
Appendix E & \hyperref[sec:AppendixE]{MLLM Templates} \\
Appendix F & \hyperref[sec:AppendixF]{More Qualitative Showcases} \\
Appendix G & \hyperref[sec:AppendixG]{Discussions} \\
\end{tabular}
\clearpage

\section*{Appendix A More Quantitative Analysis}
\label{sec:AppendixA}
\newcommand{\A}{A}
\renewcommand{\thefigure}{\A\arabic{figure}}
\renewcommand{\thetable}{\A\arabic{table}}
\renewcommand{\thesubsection}{\arabic{subsection}}
\setcounter{page}{1}
\setcounter{subsection}{0}
\setcounter{figure}{0}
\setcounter{table}{0}

\subsection*{Inference Cost Analysis}
\begin{figure}[!h]
  \centering
  \includegraphics[height=4cm]{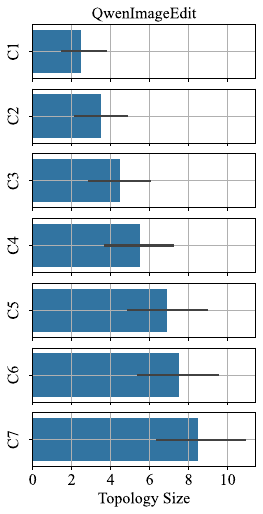}
  \includegraphics[height=4cm]{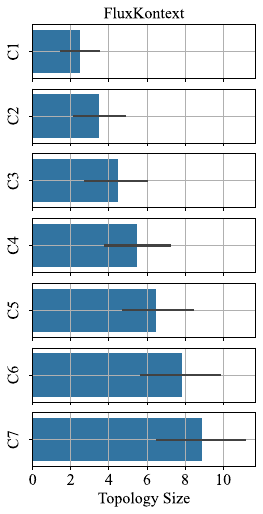}
  \includegraphics[height=4cm]{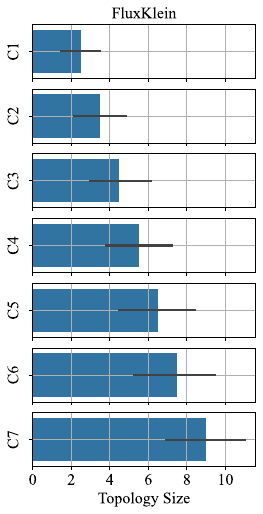}
  \caption{Inference Topology Size Across Complexities.}
  \label{fig:TopologySize}
\end{figure}

We give statistics of the inference cost by aggregating the size of the inference topology when the final optimal state is found. Topology size is defined as the number of non-root states. The results are shown as \Cref{fig:TopologySize} and \Cref{table:TopologySize}.

\begin{table}[!h]
  \caption{Average Topology Size \vs Complexity \& Linear Model Fitting. 
}
  \label{table:TopologySize}
  \fontsize{6.5pt}{6.5pt}\selectfont
  \centering
\begin{tabular}{@{}ccc|cc|cc@{}}
\toprule
\multicolumn{7}{c}{\textbf{Topology Size Statistics}} \\
\midrule
\multirow{2}{*}{Complexity} & \multicolumn{2}{c|}{QwenImageEdit}                                                                                & \multicolumn{2}{c|}{FluxKontext}                                                                                  & \multicolumn{2}{c}{FluxKlein}                                                                                     \\ \cmidrule(l){2-7} 
                            & \begin{tabular}[c]{@{}c@{}}Average\\ Size\end{tabular} & \begin{tabular}[c]{@{}c@{}}Std\\ Err\end{tabular} & \begin{tabular}[c]{@{}c@{}}Average\\ Size\end{tabular} & \begin{tabular}[c]{@{}c@{}}Std\\ Err\end{tabular} & \begin{tabular}[c]{@{}c@{}}Average\\ Size\end{tabular} & \begin{tabular}[c]{@{}c@{}}Std\\ Err\end{tabular} \\ \midrule
\multicolumn{1}{c|}{1}      & 2.39                                                   & 1.09                                                     & 2.17                                                   & 1.05                                                     & 2.12                                                   & 1.02                                                     \\
\multicolumn{1}{c|}{2}      & 3.84                                                   & 1.53                                                     & 3.65                                                   & 1.49                                                     & 3.69                                                   & 1.59                                                     \\
\multicolumn{1}{c|}{3}      & 4.76                                                   & 2.04                                                     & 4.70                                                   & 2.09                                                     & 4.49                                                   & 2.00                                                     \\
\multicolumn{1}{c|}{4}      & 6.51                                                   & 2.50                                                     & 6.24                                                   & 2.48                                                     & 6.32                                                   & 2.38                                                     \\
\multicolumn{1}{c|}{5}      & 7.33                                                   & 3.07                                                     & 7.28                                                   & 3.08                                                     & 7.23                                                   & 2.95                                                     \\
\multicolumn{1}{c|}{6}      & 8.89                                                   & 3.27                                                     & 8.82                                                   & 3.36                                                     & 9.03                                                   & 3.26                                                     \\
\multicolumn{1}{c|}{7}      & 10.04                                                  & 3.93                                                     & 9.71                                                   & 3.88                                                     & 9.95                                                   & 3.89                                                     \\ 
\end{tabular}
\\
\begin{tabular}{@{}c|ccccccc@{}}
\toprule
\multicolumn{8}{c}{\textbf{Model Fitting}} \\
\midrule
\multirow{3}{*}{\begin{tabular}[c]{@{}c@{}}Qwen\\ Image\\ Edit\end{tabular}} & Coef  & \begin{tabular}[c]{@{}c@{}}Fit\\ Value\end{tabular} & \begin{tabular}[c]{@{}c@{}}Std\\ Err\end{tabular} & t      & p-value & 2.5\% LB & 97.5\% UB \\
                                                                             & slope & 1.2721                                              & 0.036                                             & 35.306 & 0.000   & 1.180   & 1.365    \\
                                                                             & bias  & 1.1629                                              & 0.161                                             & 7.217  & 0.001   & 0.749   & 1.577    \\ \midrule
\multirow{3}{*}{\begin{tabular}[c]{@{}c@{}}Flux\\ Kontext\end{tabular}}      & Coef  & \begin{tabular}[c]{@{}c@{}}Fit\\ Value\end{tabular} & \begin{tabular}[c]{@{}c@{}}Std\\ Err\end{tabular} & t      & p-value & 2.5\% LB & 97.5\% UB \\
                                                                             & slope & 1.2693                                              & 0.031                                             & 40.565 & 0.000   & 1.189   & 1.350    \\
                                                                             & bias  & 1.0043                                              & 0.140                                             & 7.177  & 0.001   & 0.645   & 1.364    \\ \midrule
\multirow{3}{*}{\begin{tabular}[c]{@{}c@{}}Flux\\ Klein\end{tabular}}        & Coef  & \begin{tabular}[c]{@{}c@{}}Fit\\ Value\end{tabular} & \begin{tabular}[c]{@{}c@{}}Std\\ Err\end{tabular} & t      & p-value & 2.5\% LB & 97.5\% UB \\
                                                                             & slope & 1.3182                                              & 0.047                                             & 27.764 & 0.000   & 1.196   & 1.440    \\
                                                                             & bias  & 0.8457                                              & 0.212                                             & 3.983  & 0.010   & 0.300   & 1.392    \\ \bottomrule
\end{tabular}
\end{table}

To explore the time complexity, we fit linear models as in \cref{table:TopologySize}. The results suggest that MSRAMIE has an empirical linear time complexity between $O(1.18*C)$ and $O(1.44*C)$ as the instruction complexity increases.

\clearpage

\section*{Appendix B Inference-scaling Study}
\label{sec:AppendixB}
\newcommand{\B}{B}
\renewcommand{\thefigure}{\B\arabic{figure}}
\renewcommand{\thetable}{\B\arabic{table}}
\renewcommand{\thesubsection}{\arabic{subsection}}
\setcounter{subsection}{0}
\setcounter{figure}{0}
\setcounter{table}{0}
In this Appendix, we present the quantitative analysis of the effects of scaling inference steps and inference depth.
\subsection{Inference Steps Scaling} Because our method constructs a depth-first expanded inference topology, the impact of the inference steps can be examined by restricting the topology to the first N steps. We evaluated all four metrics of the optimal solutions found across varying complexity and numbers of carried inference steps. 

\begin{figure}[]
  \centering
  \includegraphics[height=5.5cm]{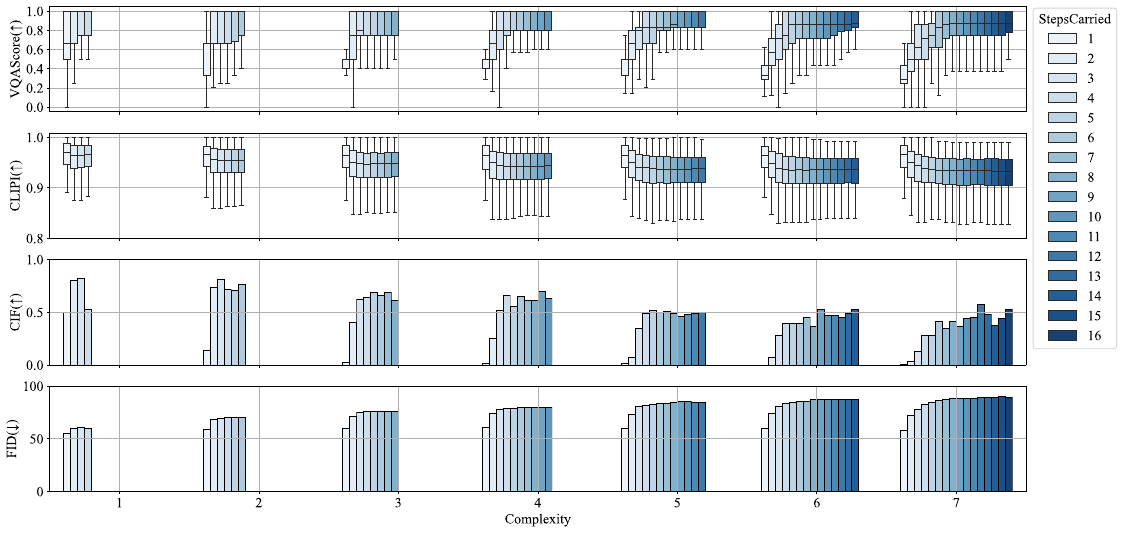}
  \caption{Performance \vs Inference Steps. (1) As complexity increases, all the four metrics degrades. (2) On the same level of complexity, as the number of carried steps increases, Instruction Following metrics are greatly improved while Identity Preservation and Perceptual Quality metrics degrade, reflecting a trade-off among these three dimensions.
  }
  \label{fig:QuantNsteps}
\end{figure}

As shown in \Cref{fig:QuantNsteps}, increasing complexity degrades both instruction following (IF) and identity preservation (IP). However, allowing more inference steps improves the average quality of candidate solutions. VQAScore and CIF increase gradually with additional steps, whereas CLIP-I tends to decrease, reflecting the trade-off between IF and IP. In contrast, Perceptual quality (PQ) consistently declines with both increasing complexity and step count, as the result of quality loss through repeated image encoding and decoding. These results highlight the need to balance improved IF against maintaining strong IP and PQ when setting the optimal number of inference steps.

\subsection{Inference Depth Scaling} 
\begin{figure}[]
  \centering
  \includegraphics[height=4.5cm]{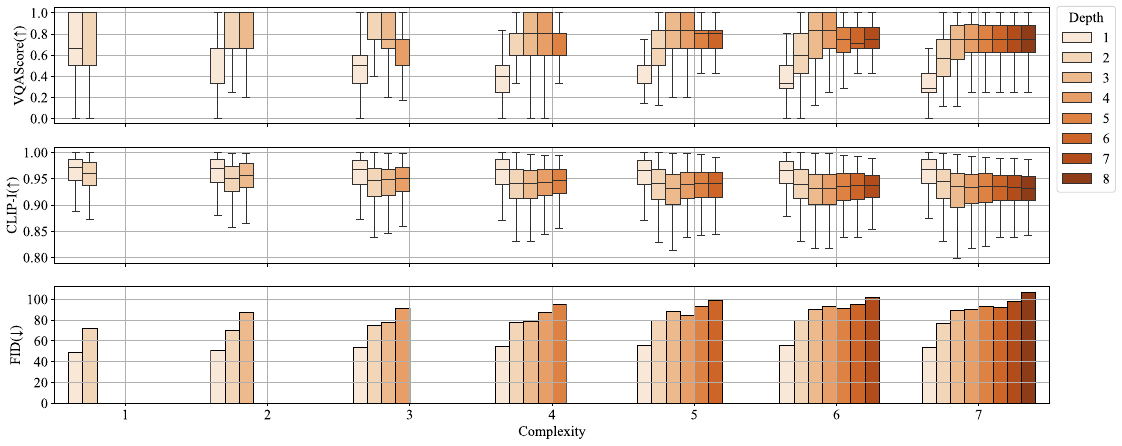}
  \caption{Performance \vs Inference Depth. (1) As complexity increases, all the four metrics degrades. (2) As the inference depth increases, Instruction Following is improved, Identity Preservation and Perceptual Quality degrades because more edited contents deviate the result further from the original image, and repeated image encoding and decoding lead to the loss of details.}
  \label{fig:QuantDepth}
\end{figure}

We performed a parallel analysis of the inference depth. The states collected at different depths are aggregated for analysis. As shown in \Cref{fig:QuantDepth}, increasing depth and task complexity produces trends consistent with those observed for inference steps.
While increased complexity impairs all evaluation dimensions, greater depth enables the discovery of solutions with improved Instruction Following (IF), indicating that deeper exploration of the inference topology is beneficial for satisfying complex editing requirements. However, this improvement comes at a cost: Identity Preservation (IP) and Perceptual Quality (PQ) tend to decline as depth increases, reflecting the effects of cumulative image editing,  as well as repeated image encoding and decoding operations.

\subsection{Study Summary} These analyzes highlight that both depth and step budget expand the search space and improve IF, but exacerbate the trade-off with IP and PQ. Therefore, selecting an appropriate depth constraint, in conjunction with a suitable step budget, is critical to balance the evaluation dimensions and secure a good and cost-effective result.

\clearpage

\section*{Appendix C Ablation Study Results Details}
\label{sec:AppendixC}
\newcommand{\C}{C}
\renewcommand{\thefigure}{\C\arabic{figure}}
\renewcommand{\thetable}{\C\arabic{table}}
\renewcommand{\thesubsection}{\arabic{subsection}}
\setcounter{subsection}{0}
\setcounter{figure}{0}
\setcounter{table}{0}
In this Appendix, we present the complete detail results for ablation study, as well as the explanation for the involved 5 methods.
\subsection{Method Explanation}
\textbf{Direct Editing}. The editing process performed by the Actor module is the basis of MSRAMIE ability. Direct editing tasks were launched with the base image editing model.\\
\textbf{Resampling-Only}. The resampling mechanism achieved by state branching is thought to find better solutions via best-of-N strategy. To study the effect of resampling, we directly pass the initial instruction to the Actor module rather than providing decomposed thoughts.\\
\textbf{Chain-of-States-Only (CoS-Only)}. A chain-of-states corresponds to the activated decision path within a MSRAMIE topology. Chain-of-States is regarded as the submechanism of Tree-of-States. It aggregates the partial solutions in a sequential order with each solution focusing on the drawback of the previous ones. To experiment with Chain-of-States, we limit the max number of state children to 1 and set the Retriever search range to 0.\\
\textbf{Tree-of-States-Only (ToS-Only)}. By setting Retriever search range to 0 to disable Graph-of-References, we experimented with the pure Tree-of-States mechanism. Compare with Chain-of-States, it enables the agent to backtrack to previous states and do exploration on different trajectory.\\
\textbf{MSRAMIE}. The Graph-of-References is thought to improve the efficiency of trajectory exploration by encouraging the Instruction Generator to focus on the untried items in the CheckList. By enabling it and Tree-of-States simultaneously, the full architecture of MSRAMIE is exploited.

During the test, we use different hyperparameter settings to secure a controlled inference budget for the methods involved, provided in Appendix D.

\subsection{Ablation Test Results} The detail ablation test results across all the seven complexity levels are as shown in \Cref{table:VQAScoreAblation,table:CIFAblation,table:CLIPIAblation,table:FIDAblation}. For Instruction Following, Resampling-Only, CoS-Only, ToS-Only and MSRAMIE provide progressively increased gain. At low complexity, ToS-Only is good enough, but at higher instruction complexity, enabling GoR facilitates broader solution search and provides additional gains. For Identity Preservation, complete MSRAMIE provides slightly better overall performance than ToS-Only, while CoS-Only almost has no effect. AaFor Perceptual Quality, Resampling-only yields modest gains. In contrast, CoS-only performs poorly because single-sample generation has high variance, leading to accumulated degradation. ToS-only mitigates this issue and improves overall performance.  We also observe a lower FID for resampling-only, which results from limited visual change caused by insufficient editing. As inference depth increases in CoS-only, ToS-only, and full MSRAMIE, FID rises with the extent of image modification; however, it remains lower than direct editing for comparable editing levels.
\begin{table}[]
\fontsize{6.5pt}{6.5pt}\selectfont
\centering
\caption{VQAScore Ablation Test Results.}
\label{table:VQAScoreAblation}
\begin{tabular}{@{}l|ccccc|@{}}
\toprule
\multicolumn{6}{c}{\textbf{VQAScore($\uparrow$)}} \\ 
\midrule
\multicolumn{1}{|l|}{\textbf{Complexity}}                & \multicolumn{1}{c}{Direct} & \multicolumn{1}{c}{Resampling-Only} & \multicolumn{1}{c}{CoS-Only} & \multicolumn{1}{c}{ToS-Only} & MSRAMIE        \\ \midrule
\multicolumn{1}{|l|}{1}                         & 0.8088                               & 0.8514                                                                & 0.8396                                                         & {\ul \textbf{0.8836}}                  & 0.8760                 \\
\multicolumn{1}{|l|}{2}                         & 0.8041                               & 0.8283                                                                & 0.8449                                                         & 0.8750                                 & {\ul \textbf{0.8947}}  \\
\multicolumn{1}{|l|}{3}                         & 0.7795                               & 0.8037                                                                & 0.8280                                                         & 0.8738                                 & {\ul \textbf{0.8854}}  \\
\multicolumn{1}{|l|}{4}                         & 0.7530                               & 0.7794                                                                & 0.8000                                                         & 0.8674                                 & {\ul \textbf{0.8768}}  \\
\multicolumn{1}{|l|}{5}                         & 0.7412                               & 0.7706                                                                & 0.8183                                                         & 0.8578                                 & {\ul \textbf{0.8870}}  \\
\multicolumn{1}{|l|}{6}                         & 0.7325                               & 0.7635                                                                & 0.7880                                                         & 0.8566                                 & {\ul \textbf{0.8756}}  \\
\multicolumn{1}{|l|}{7}                         & 0.7172                               & 0.7449                                                                & 0.7615                                                         & 0.8520                                 & {\ul \textbf{0.8780}}  \\ \midrule
\multicolumn{1}{|l|}{\textbf{Overall}}                   & 0.7623                               & 0.7917                                                                & 0.8115                                                         & 0.8666                                 & {\ul \textbf{0.8819}}  \\ \midrule
\multicolumn{1}{|l|}{\textbf{Complexity}}                & \multicolumn{5}{c|}{\textbf{Improvement}}                                                                                                                                                                                                         \\ \midrule
\multicolumn{1}{|l|}{1}                         & -            & 5.26\%                                        & 3.81\%                                 & {\ul \textbf{9.24\%}}                  & 8.31\%                 \\
\multicolumn{1}{|l|}{2}                         & -                                    & 3.02\%                                                                & 5.08\%                                                         & 8.83\%                                 & {\ul \textbf{11.27\%}} \\
\multicolumn{1}{|l|}{3}                         & -                                    & 3.10\%                                                                & 6.23\%                                                         & 12.10\%                                & {\ul \textbf{13.59\%}} \\
\multicolumn{1}{|l|}{4}                         & -                                    & 3.51\%                                                                & 6.24\%                                                         & 15.20\%                                & {\ul \textbf{16.44\%}} \\
\multicolumn{1}{|l|}{5}                         & -                                    & 3.96\%                                                                & 10.40\%                                                        & 15.72\%                                & {\ul \textbf{19.67\%}} \\
\multicolumn{1}{|l|}{6}                         & -                                    & 4.23\%                                                                & 7.57\%                                                         & 16.94\%                                & {\ul \textbf{19.53\%}} \\
\multicolumn{1}{|l|}{7}                         & -                                    & 3.86\%                                                                & 6.18\%                                                         & 18.79\%                                & {\ul \textbf{22.41\%}} \\ \midrule
\multicolumn{1}{|l|}{\textbf{Overall}}                   & -                                    & 3.85\%                                                                & 6.50\%                                                         & 13.83\%                                & {\ul \textbf{15.89\%}} \\ \bottomrule
\end{tabular}
\end{table}

\begin{table}[]
\fontsize{6.5pt}{6.5pt}\selectfont
\centering
\caption{CIF Ablation Test Results.}
\label{table:CIFAblation}
\begin{tabular}{@{}l|ccccc|@{}}
\toprule
\multicolumn{6}{c}{\textbf{CIFScore($\uparrow$)}} \\ 
\midrule
\multicolumn{1}{|l|}{\textbf{Complexity}}                & \multicolumn{1}{c}{Direct} & \multicolumn{1}{c}{Resampling-Only} & \multicolumn{1}{c}{CoS-Only} & \multicolumn{1}{c}{ToS-Only} & MSRAMIE        \\ \midrule
\multicolumn{1}{|l|}{1}                         & 0.6064                               & 0.6930                                                           & 0.6855                                                    & {\ul \textbf{0.7382}}             & 0.7307                  \\
\multicolumn{1}{|l|}{2}                         & 0.5292                               & 0.5819                                                           & 0.6554                                                    & 0.7100                            & {\ul \textbf{0.7307}}   \\
\multicolumn{1}{|l|}{3}                         & 0.4011                               & 0.4407                                                           & 0.5687                                                    & 0.6441                            & {\ul \textbf{0.6478}}   \\
\multicolumn{1}{|l|}{4}                         & 0.3032                               & 0.3503                                                           & 0.4769                                                    & 0.5857                            & {\ul \textbf{0.6008}}   \\
\multicolumn{1}{|l|}{5}                         & 0.2241                               & 0.2976                                                           & 0.4444                                                    & 0.5141                            & {\ul \textbf{0.5405}}   \\
\multicolumn{1}{|l|}{6}                         & 0.2166                               & 0.2637                                                           & 0.4011                                                    & 0.4821                            & {\ul \textbf{0.4840}}   \\
\multicolumn{1}{|l|}{7}                         & 0.1676                               & 0.1977                                                           & 0.3070                                                    & 0.4520                            & {\ul \textbf{0.4708}}   \\ \midrule
\multicolumn{1}{|l|}{\textbf{Overall}}                   & 0.3497                               & 0.4036                                                           & 0.5056                                                    & 0.5895                            & {\ul \textbf{0.6008}}   \\ \midrule
\multicolumn{1}{|l|}{\textbf{Complexity}}                & \multicolumn{5}{c|}{\textbf{Improvement}}                                                                                                                                                                                           \\ \midrule            
\multicolumn{1}{|l|}{1}                         & -         & 14.29\%                  & 13.04\%                           & {\ul \textbf{21.74\%}}            & 20.50\%                 \\
\multicolumn{1}{|l|}{2}                         & -                                    & 9.96\%                                                           & 23.84\%                                                   & 34.16\%                           & {\ul \textbf{38.08\%}}  \\
\multicolumn{1}{|l|}{3}                         & -                                    & 9.86\%                                                           & 41.78\%                                                   & 60.56\%                           & {\ul \textbf{61.50\%}}  \\
\multicolumn{1}{|l|}{4}                         & -                                    & 15.53\%                                                          & 57.27\%                                                   & 93.17\%                           & {\ul \textbf{98.14\%}}  \\
\multicolumn{1}{|l|}{5}                         & -                                    & 32.77\%                                                          & 98.32\%                                                   & 129.41\%                          & {\ul \textbf{141.18\%}} \\
\multicolumn{1}{|l|}{6}                         & -                                    & 21.74\%                                                          & 85.22\%                                                   & 122.61\%                          & {\ul \textbf{123.48\%}} \\
\multicolumn{1}{|l|}{7}                         & -                                    & 17.98\%                                                          & 83.15\%                                                   & 169.66\%                          & {\ul \textbf{180.90\%}} \\ \midrule
\multicolumn{1}{|l|}{\textbf{Overall}}                   & -                                    & 17.45\%                                                          & 57.52\%                                                   & 90.19\%                           & {\ul \textbf{94.82\%}}  \\ \bottomrule
\end{tabular}
\end{table}

\begin{table}[]
\caption{CLIP-I Ablation Test Results.}
\label{table:CLIPIAblation}
\fontsize{6.5pt}{6.5pt}\selectfont
\centering
\begin{tabular}{lcccccccc}
\toprule
\multicolumn{9}{c}{\textbf{CLIP-I($\uparrow$)}}                                                                                                                                                                                                                                              \\ \midrule
\multicolumn{1}{|l|}{\textbf{Complexity}} & Direct & \begin{tabular}[c]{@{}c@{}}Resampling-Only\end{tabular} & Improvement & \multicolumn{1}{c|}{P-value}  & Direct & \begin{tabular}[c]{@{}c@{}}CoS-Only\end{tabular} & Improvement & \multicolumn{1}{c|}{P-value}                 \\ \midrule
\multicolumn{1}{|l|}{1}          & 0.9542 & 0.9562                                                     & 0.20\%      & \multicolumn{1}{c|}{2.62E-09} & 0.9524 & 0.9536                                              & 0.12\%      & \multicolumn{1}{c|}{8.15E-02} \\
\multicolumn{1}{|l|}{2}          & 0.9441 & 0.9463                                                     & 0.24\%      & \multicolumn{1}{c|}{3.02E-09} & 0.9420 & 0.9421                                              & 0.02\%      & \multicolumn{1}{c|}{8.63E-01}                \\
\multicolumn{1}{|l|}{3}          & 0.9389 & 0.9409                                                     & 0.22\%      & \multicolumn{1}{c|}{2.24E-07} & 0.9381 & 0.9362                                              & -0.20\%     & \multicolumn{1}{c|}{6.16E-02} \\
\multicolumn{1}{|l|}{4}          & 0.9324 & 0.9341                                                     & 0.18\%      & \multicolumn{1}{c|}{3.90E-04} & 0.9317 & 0.9299                                              & -0.19\%     & \multicolumn{1}{c|}{1.56E-01}                \\
\multicolumn{1}{|l|}{5}          & 0.9318 & 0.9336                                                     & 0.19\%      & \multicolumn{1}{c|}{1.44E-07} & 0.9261 & 0.9266                                              & 0.06\%      & \multicolumn{1}{c|}{7.12E-01}                \\
\multicolumn{1}{|l|}{6}          & 0.9287 & 0.9299                                                     & 0.13\%      & \multicolumn{1}{c|}{1.51E-03} & 0.9165 & 0.9163                                              & -0.03\%     & \multicolumn{1}{c|}{8.68E-01}                \\
\multicolumn{1}{|l|}{7}          & 0.9224 & 0.9243                                                     & 0.20\%      & \multicolumn{1}{c|}{1.22E-06} & 0.9172 & 0.9192                                              & 0.22\%      & \multicolumn{1}{c|}{1.87E-01}                \\ \midrule
\multicolumn{1}{|l|}{\textbf{Overall}}                          & 0.9370 & 0.9388                                                     & 0.20\%      & \multicolumn{1}{l|}{-}          & 0.9355 & 0.9354                                              & 0.00\%      & \multicolumn{1}{l|}{-}                         \\ \hline

                                                                                                                                                               \\ \midrule
\multicolumn{1}{|l|}{\textbf{Complexity}} & Direct & ToS-Only & Improvement           & \multicolumn{1}{c|}{P-value}  & Direct & MSRAMIE & Improvement           & \multicolumn{1}{c|}{P-value}                 \\ \midrule
\multicolumn{1}{|l|}{1}          & 0.9530 & 0.9557   & 0.28\%                & \multicolumn{1}{c|}{1.14E-04} & 0.9532 & 0.9575  & {\ul \textbf{0.45\%}} & \multicolumn{1}{c|}{2.40E-10} \\
\multicolumn{1}{|l|}{2}          & 0.9416 & 0.9448   & 0.34\%                & \multicolumn{1}{c|}{2.81E-03} & 0.9410 & 0.9446  & {\ul \textbf{0.38\%}} & \multicolumn{1}{c|}{4.43E-04}                \\
\multicolumn{1}{|l|}{3}          & 0.9333 & 0.9368   & {\ul \textbf{0.37\%}} & \multicolumn{1}{c|}{1.21E-03} & 0.9356 & 0.9380  & 0.26\%                & \multicolumn{1}{c|}{2.49E-02} \\
\multicolumn{1}{|l|}{4}          & 0.9283 & 0.9312   & 0.31\%                & \multicolumn{1}{c|}{2.41E-02} & 0.9289 & 0.9322  & {\ul \textbf{0.36\%}} & \multicolumn{1}{c|}{9.76E-03}                \\
\multicolumn{1}{|l|}{5}          & 0.9231 & 0.9253   & 0.23\%                & \multicolumn{1}{c|}{1.32E-01} & 0.9221 & 0.9249  & {\ul \textbf{0.31\%}} & \multicolumn{1}{c|}{1.43E-02}                \\
\multicolumn{1}{|l|}{6}          & 0.9118 & 0.9154   & {\ul \textbf{0.39\%}} & \multicolumn{1}{c|}{2.16E-02} & 0.9146 & 0.9181  & 0.38\%                & \multicolumn{1}{c|}{2.35E-02}                \\
\multicolumn{1}{|l|}{7}          & 0.9082 & 0.9140   & 0.64\%                & \multicolumn{1}{c|}{1.96E-03} & 0.9106 & 0.9190  & {\ul \textbf{0.91\%}} & \multicolumn{1}{c|}{2.82E-07}                \\ \midrule
\multicolumn{1}{|l|}{\textbf{Overall}}                          & 0.9333 & 0.9365   & 0.35\%                & \multicolumn{1}{l|}{-}           & 0.9338 & 0.9376  & {\ul \textbf{0.41\%}} & \multicolumn{1}{l|}{-}                                              \\ \bottomrule
\end{tabular}
\end{table}

\begin{table}[]
\caption{FID Ablation Test Results. P-value is not computed as for CLIP-I, because FID is not computed image-wisely but between the original and edited image sets.}
\label{table:FIDAblation}
\fontsize{6.5pt}{6.5pt}\selectfont
\centering
\begin{tabular}{@{}lcccccc@{}}
\toprule
\multicolumn{7}{c}{\textbf{FID($\downarrow$)}}                                                                                                                                                              \\ \midrule
\multicolumn{1}{|l|}{\textbf{Complexity}} & Direct & Resampling-Only & \multicolumn{1}{c|}{Improvement}            & \multicolumn{1}{l}{Direct} & CoS-Only & \multicolumn{1}{c|}{Improvement} \\ \midrule
\multicolumn{1}{|l|}{1}                   & 72.56  & 72.33           & \multicolumn{1}{c|}{0.32\%}                 & 76.86                      & 76.71    & \multicolumn{1}{c|}{0.20\%}      \\
\multicolumn{1}{|l|}{2}                   & 80.09  & 80.01           & \multicolumn{1}{c|}{0.10\%}                 & 88.37                & 89.92    & \multicolumn{1}{c|}{-1.75\%}     \\
\multicolumn{1}{|l|}{3}                   & 88.58 & 87.57          & \multicolumn{1}{c|}{{\ul \textbf{1.14\%}}}  & 100.31                     & 105.25   & \multicolumn{1}{c|}{-4.92\%}     \\
\multicolumn{1}{|l|}{4}                   & 95.34 & 96.12          & \multicolumn{1}{c|}{-0.82\%}                & 127.35               & 132.21   & \multicolumn{1}{c|}{-3.82\%}     \\
\multicolumn{1}{|l|}{5}                   & 99.51 & 97.09          & \multicolumn{1}{c|}{{\ul \textbf{2.43\%}}}  & 132.91                     & 135.53   & \multicolumn{1}{c|}{-1.97\%}     \\
\multicolumn{1}{|l|}{6}                   & 101.26 & 101.44          & \multicolumn{1}{c|}{{\ul \textbf{-0.18\%}}} & 135.06                     & 141.38   & \multicolumn{1}{c|}{-4.68\%}     \\
\multicolumn{1}{|l|}{7}                   & 110.12 & 110.19          & \multicolumn{1}{c|}{-0.06\%}                & 141.75                     & 143.99   & \multicolumn{1}{c|}{-1.58\%}     \\ \midrule
\multicolumn{1}{|l|}{\textbf{Overall}}    & 91.27 & 90.89          & \multicolumn{1}{c|}{{\ul \textbf{0.42\%}}}  & 107.54                     & 110.38   & \multicolumn{1}{c|}{-2.64\%}        \\ \hline

                                                                                                                                                               \\ \midrule
\multicolumn{1}{|l|}{\textbf{Complexity}} & Direct & ToS-Only & \multicolumn{1}{c|}{Improvement}           & \multicolumn{1}{l}{Direct} & MSRAMIE & \multicolumn{1}{c|}{Improvement}           \\ \midrule
\multicolumn{1}{|l|}{1}                   & 75.54  & 73.79    & \multicolumn{1}{c|}{2.32\%}                & 75.61                      & 72.48   & \multicolumn{1}{c|}{{\ul \textbf{4.14\%}}} \\
\multicolumn{1}{|l|}{2}                   & 87.37  & 88.05    & \multicolumn{1}{c|}{-0.78\%}               & 88.38                & 88.18   & \multicolumn{1}{c|}{{\ul \textbf{0.23\%}}} \\
\multicolumn{1}{|l|}{3}                   & 103.59 & 103.62   & \multicolumn{1}{c|}{-0.03\%}      & 98.49                      & 99.73   & \multicolumn{1}{c|}{-1.26\%}               \\
\multicolumn{1}{|l|}{4}                   & 121.77 & 119.50   & \multicolumn{1}{c|}{{\ul \textbf{1.86\%}}} & 121.70               & 121.85  & \multicolumn{1}{c|}{-0.12\%}               \\
\multicolumn{1}{|l|}{5}                   & 136.20 & 137.78   & \multicolumn{1}{c|}{-1.16\%}      & 141.18                     & 139.87  & \multicolumn{1}{c|}{0.93\%}                \\
\multicolumn{1}{|l|}{6}                   & 144.36 & 148.40   & \multicolumn{1}{c|}{-2.80\%}      & 140.13                     & 146.20  & \multicolumn{1}{c|}{-4.33\%}               \\
\multicolumn{1}{|l|}{7}                   & 153.30 & 159.78   & \multicolumn{1}{c|}{-4.23\%}               & 150.05                     & 148.35  & \multicolumn{1}{c|}{{\ul \textbf{1.13\%}}} \\ \midrule
\multicolumn{1}{|l|}{\textbf{Overall}}    & 108.24 & 108.81   & \multicolumn{1}{c|}{-0.52\%}      & 107.91                     & 107.78  & \multicolumn{1}{c|}{{\ul \textbf{0.12\%}}} \\ \bottomrule
\end{tabular}
\end{table}

\clearpage

\section*{Appendix D Experiment Setup}
\label{sec:AppendixD}
\newcommand{\D}{D}
\renewcommand{\thefigure}{\D\arabic{figure}}
\renewcommand{\thetable}{\D\arabic{table}}
\renewcommand{\thesubsection}{\arabic{subsection}}
\setcounter{subsection}{0}
\setcounter{figure}{0}
\setcounter{table}{0}
In this Appendix, we present the hyperparameter settings for MSRAMIE in the quantitative study and ablation study. The explanation for each hyperparameter is also provided.
\subsection{Hyperparameters}
\subsubsection{General} These hyperparameters are for the general control of MSRAMIE: 
\begin{itemize}
    \item MaxSteps: The maximum number of inference steps, also the number of states in the topology excluding the root. It defines the overall inference budget.
\end{itemize}
\subsubsection{Instructor} These hyperparameters are for the general control of the top-level module Instructor:
\begin{itemize}
    \item MLLM: The multimodal large language model used as the backend for instruction generation.
    \item MaxNTry: The maximum number of retries for instruction generation to handle potential formatting errors.
    \item InstructionVolume: The maximum number of sub-instructions contained in the generated state thought.
\end{itemize}
\subsubsection{Actor} These hyperparameters are for the control of the top-level module Actor:
\begin{itemize}
    \item ImageEditingPipeline: The instruction-based image editing model used as the Actor.
\end{itemize}
\subsubsection{Retriever} These hyperparameters are for the control of the submodule Retriever:
\begin{itemize}
    \item SearchRange: Limitation of the neighbourhood in the topology for retrieving reference states (0 disables retrieval, 1 searches states between the adjacent two layers, 2 searches states between the adjacent four layers, etc.).
    \item TopK: The maximum number of reference states selected.
    \item RelevanceThreshold: An integer between 0-100 which sets the minimum similarity required for candidate reference states.
\end{itemize}

\subsubsection{Evaluator} These hyperparameters are for the control of the submodule Evaluator:
\begin{itemize}
    \item MLLM: The multimodal model used as the backend for state evaluation.
\end{itemize}
\subsubsection{Scheduler} These hyperparameters are for the control of the submodule Scheduler:
\begin{itemize}
    \item MaxNChildren: Max number of children per state in the inference topology.
    \item MaxDepth: Max depth of states in the inference topology. Root state has depth 0.
    \item MinDepth: A minimum depth for final solutions, preventing shallow states from being selected even if they achieve higher scores.
    \item CompletionThreshold: The minimum metric values required to terminate inference.
    \item DegradeTolerance: The maximum allowable performance drop along a decision path. Exceeding this threshold triggers pruning and backtracking.
    \item StayThreshold: The minimum metric values required for a state to be eligible as the next parent state.
\end{itemize}

\subsection{Quantitative Study Settings} The parameter settings in \Cref{table:MainTestParams} are applied for quantitative study. We apply the same setting across all tested models to avoid any influence of the potential hyperparameter sensitivity. 

\begin{table}[]
\caption{Quantitative Study Hyperparameter Settings. C stands for instruction complexity, which ranges from 1 to 7.
}
\label{table:MainTestParams}
\centering
\begin{tabular}{@{}lll@{}}
\toprule
Modules                      & Hyperparams                      & Value                                      \\ \midrule
General                      & MaxSteps                         & 2*(C+1)                                    \\ \midrule
                             & MLLM                             & Qwen3VL                                    \\
                             & MaxNTry                          & 3                                          \\
\multirow{-3}{*}{Instructor} & InstructionVolume                & 2                                          \\ \midrule
Actor                        & ImageEditingPipeline             & AnyModel                                  \\ \midrule
                             & SearchRange                      & 2                                          \\
                             & TopK                             & 3                                          \\
\multirow{-3}{*}{Retriever}  & RelevanceThreshold               & 50                                         \\ \midrule
Evaluator                    & MLLM                             & Qwen3VL                                    \\ \midrule
                             & MaxNChildren                     & 2                                          \\
                             & MaxDepth                         & C+1                                        \\
                             & MinDepth                         & ((C+1)+(C+1)\%2)//2 \\
                             & CompletionThreshold              & VQAScore-1,   CLIPI-1                      \\
                             & DegradeTolerance                 & VQAScore-0                                 \\
\multirow{-6}{*}{Scheduler}  & StayThreshold                    & VQAScore-0.1                               \\ \bottomrule
\end{tabular}
\end{table}

\subsection{Ablation Study Settings} The parameter settings in \Cref{table:AblationParams} are applied for the ablation test.
\label{sec:ablation_params}
\begin{table}[]
\caption{Ablation Study Hyperparameter Settings. C stands for instruction complexity, which ranges from 1 to 7. Key hyperparameters that differentiate involved methods are colored. 
}
\label{table:AblationParams}
\centering
\begin{tabular}{@{}llll@{}}
\toprule
Modules                      & Hyperparams                      & \textbf{Resampling-Only}                    & \textbf{CoS-Only}                          \\ \midrule
General                      & MaxSteps                         & 2*(C+1)                                     & 2*(C+1)                                    \\ \midrule
                             & MLLM                             & -                                           & Qwen3VL                                    \\
                             & MaxNTry                          & -                                           & 3                                          \\
\multirow{-3}{*}{Instructor} & InstructionVolume                & -                                           & 2                                          \\ \midrule
Actor                        & ImageEditingPipeline             & QwenImageEdit                               & QwenImageEdit                              \\ \midrule
                             & SearchRange                      & \textcolor{red}{0}                          & \textcolor{red}{0}                         \\
                             & TopK                             & 3                                           & 3                                          \\
\multirow{-3}{*}{Retriever}  & RelevanceThreshold               & 50                                          & 50                                         \\ \midrule
Evaluator                    & MLLM                             & Qwen3VL                                     & Qwen3VL                                    \\ \midrule
                             & MaxNChildren                     & \textcolor{blue}{2*(C+1)}                   & \textcolor{blue}{1}                        \\
                             & MaxDepth                         & \textcolor{green}{1}                        & \textcolor{green}{2*(C+1)}                       \\
                             & MinDepth                         & 1                                           & ((C+1)+(C+1)\%2)//2                        \\
                             & CompletionThreshold              & VQAScore-1,   CLIPI-1                       & VQAScore-1, CLIPI-1                        \\
                             & DegradeTolerance                 & VQAScore-0                                  & VQAScore-0                                 \\
\multirow{-6}{*}{Scheduler}  & StayThreshold                    & VQAScore-0.1                                & VQAScore-0.1                               \\ \bottomrule
\end{tabular}

\begin{tabular}{@{}llll@{}}
\toprule
Modules                      & Hyperparams                      & \textbf{ToS-Only}                          & \textbf{MSRAMIE}   \\ \midrule
General                      & MaxSteps                         & 2*(C+1)                                    & 2*(C+1)                                    \\ \midrule
                             & MLLM                             & Qwen3VL                                    & Qwen3VL                                    \\
                             & MaxNTry                          & 3                                          & 3                                          \\
\multirow{-3}{*}{Instructor} & InstructionVolume                & 2                                          & 2                                          \\ \midrule
Actor                        & ImageEditingPipeline             & QwenImageEdit                              & QwenImageEdit                              \\ \midrule
                             & SearchRange                      & \textcolor{red}{0}                         & \textcolor{red}{2}                         \\
                             & TopK                             & 3                                          & 3                                          \\
\multirow{-3}{*}{Retriever}  & RelevanceThreshold               & 50                                         & 50                                         \\ \midrule
Evaluator                    & MLLM                             & Qwen3VL                                    & Qwen3VL                                    \\ \midrule
                             & MaxNChildren                     & \textcolor{blue}{2}                        & \textcolor{blue}{2}                        \\
                             & MaxDepth                         & \textcolor{green}{C}                       & \textcolor{green}{C}                       \\
                             & MinDepth & ((C+1)+(C+1)\%2)//2   & ((C+1)+(C+1)\%2)//2                        \\
                             & CompletionThreshold              & VQAScore-1,   CLIPI-1                      & VQAScore-1, CLIPI-1                        \\
                             & DegradeTolerance                 & VQAScore-0                                 & VQAScore-0                                 \\
\multirow{-6}{*}{Scheduler}  & StayThreshold                    & VQAScore-0.1                               & VQAScore-0.1                               \\ \bottomrule
\end{tabular}
\end{table}
\clearpage

\section*{Appendix E MLLM Templates}
\label{sec:AppendixE}
\newcommand{\E}{E}
\renewcommand{\thefigure}{\E\arabic{figure}}
\renewcommand{\thetable}{\E\arabic{table}}
\renewcommand{\thesubsection}{\arabic{subsection}}
\setcounter{subsection}{0}
\setcounter{figure}{0}
\setcounter{table}{0}
In this Appendix, the MLLM prompt templates used in the Instructor submodules Instruction Generator and Evaluator are presented. In the templates, we color the contextual variables with \textcolor{Blue}{Blue} to label them from the fixed content, and explain or exemplify them later. Besides, we present the regular expressions used for guided decoding, which secures a formatted generation of the MLLM.
\subsection{Instructiong Generator}
\subsubsection{System Prompt} The following system prompt template is used.
\begin{lstlisting}[basicstyle=\ttfamily\small, breaklines=true]
You are an image-editing work instructor. The user wishes to edit an image to meet some editing requirements. A requirement checklist is provided. Your task is to generate an instruction that only do up to (*@\textcolor{Blue}{IV}@*) modifications to the "latest_image". Even it cannot fullfill all the requirements.
\end{lstlisting}

\subsubsection{User Prompt} The following system prompt template is used.
\begin{lstlisting}[basicstyle=\ttfamily\small, breaklines=true]
{'Requirement Checklist': (*@\textcolor{Blue}{CHECKLIST}@*),
'References': {
{'Reference_0': (*@\textcolor{Blue}{THOUGHT}@*), 'CaseSimilarity': (*@\textcolor{Blue}{SIMSCORE}@*)},
{'Reference_1': (*@\textcolor{Blue}{THOUGHT}@*), 'CaseSimilarity': (*@\textcolor{Blue}{SIMSCORE}@*)},
...
}
\end{lstlisting}

\subsubsection{Variable Explanation}
\begin{itemize}
    \item \textcolor{Blue}{IV}: The InstructionVolume hyperparameter. Please refer to Appendix D for details.
    \item \textcolor{Blue}{CHECKLISTS}: This is an ordered list of VQA questions with parenthesized answers (Y/N) exemplified as follow.\\
"1. Are all the existing background elements identified? (N), 2. Is the new busy metropolitan skyline image created? (N), 3. Is the current background successfully removed from the original image? (N), 4. Is the new metropolitan skyline correctly inserted into the background area? (N)"
    \item \textcolor{Blue}{THOUGHT}: A textual instruction exemplified as "Remove the current background from the image and insert the new metropolitan skyline into the background area". It comes from the reference state.
    \item \textcolor{Blue}{SIMSCORE}: An integer between 0-100.
\end{itemize}

\subsubsection{Guided-Decoding Format} The following regular expression for guided decoding is used.
\begin{lstlisting}[basicstyle=\ttfamily\small, breaklines=true]
r"""<think></think>\n?\{"reasoning":"Step1:.*?Step2:.*?Step3:.*?", ?"instruction":".*?"\}"""
\end{lstlisting}

\subsection{Evaluator-InstructionAnalyzer}
\subsubsection{System Prompt} The following system prompt template is used.
\begin{lstlisting}[basicstyle=\ttfamily\small, breaklines=true]
You are an image editing task assistant. The user will provide you with an image and an editing instruction.
Three tasks: 1.Describe the image 2.Analyze the editing instruction. Based on the original image content, decompose the instruction into smaller sub instructions. 3.Transform each sub instruction into a question to check whether they are done or not.
Strictly provide your evaluation in a dict format. For example: {"ImageDescription":"Description here.", "SubInstructions": "List sub instructions here.", "Questions": "List questions here."}.
IMPORTANT: Provide the sub instructions one by one, indexed by numbers, split by line break. e.g. 1.xxx\n 2.xxx\n ...
IMPORTANT: Provide the questions one by one, end with question mark, indexed by numbers, split by line break. e.g. 1.xxx?\n 2.xxx?\n ...
IMPORTANT: If the instruction is simple enough and cannot be decomposed even further, give "SubInstructions" that have only one item.
\end{lstlisting}

\subsubsection{User Prompt} There is no prompt template for this part. We simply feed the following two items to the MLLM: \textcolor{Blue}{INSTRUCTION} and \textcolor{Blue}{ORIGIN\_IMAGE}.

\subsubsection{Variable Explanation}
\begin{itemize}
    \item \textcolor{Blue}{INSTRUCTION}:The original multi-instruction.
    \item \textcolor{Blue}{ORIGIN\_IMAGE}: The original image.
\end{itemize}
\subsubsection{Guided-Decoding Format} The following regular expression for guided decoding is used.
\begin{lstlisting}[basicstyle=\ttfamily\small, breaklines=true]
r"""<think></think>\n?\{"ImageDescription":"The.*?", "SubInstructions":"1\..*\n.*?"\, "Questions":"1\..*\?\n.*?"\}"""
\end{lstlisting}

\subsection{Evaluator-VQAChecker} 
\subsubsection{System Prompt} The following system prompt template is used.
\begin{lstlisting}[basicstyle=\ttfamily\small, breaklines=true]
You are an image editing task assistant. The user will provide you with an original image, an edited image and a checklist of editing questions.
Your tasks: Compare the content of the two images, answer each question in the checklist. 
Strictly provide your evaluation in a dict format. For example: {"Checklist":"Output here."}.
IMPORTANT: Answer Y or N for each question, do not include extra output. e.g. 1.xxx (Y), 2.xxx (N), 3.xxx (Y), ...
\end{lstlisting}

\subsubsection{User Prompt} There is no prompt template for this part. We simply feed the following three items to the MLLM: \textcolor{Blue}{QUESTIONS}, \textcolor{Blue}{ORIGIN\_IMAGE} and \textcolor{Blue}{EDITED\_IMAGE}.
\subsubsection{Variable Explanation}
\begin{itemize}
    \item \textcolor{Blue}{QUESTIONS}:This is an ordered list of VQA questions exemplified as below,
"1. Are all the existing background elements identified? 2. Is the new busy metropolitan skyline image created? 3. Is the current background successfully removed from the original image? 4. Is the new metropolitan skyline correctly inserted into the background area? "
    \item \textcolor{Blue}{ORIGIN\_IMAGE}: The original image.
    \item \textcolor{Blue}{EDITED\_IMAGE}:  The current state output image.
\end{itemize}
\subsubsection{Guided-Decoding Format} The following regular expression for guided decoding is used.
\begin{lstlisting}[basicstyle=\ttfamily\small, breaklines=true]
r"""<think></think>\n?\{"Checklist":"(?:\d+\..*?\s\((?:Y|N)\)\n)+"\}"""
\end{lstlisting}

\clearpage

\section*{Appendix F More Qualitative Showcases}
\label{sec:AppendixF}
\newcommand{\F}{F}
\renewcommand{\thefigure}{\F\arabic{figure}}
\renewcommand{\thetable}{\F\arabic{table}}
\renewcommand{\thesubsection}{\arabic{subsection}}
\setcounter{subsection}{0}
\setcounter{figure}{0}
\setcounter{table}{0}

\subsection{Transparent Inference Process}

\begin{figure}[!h]
\centering
\includegraphics[width=11cm]{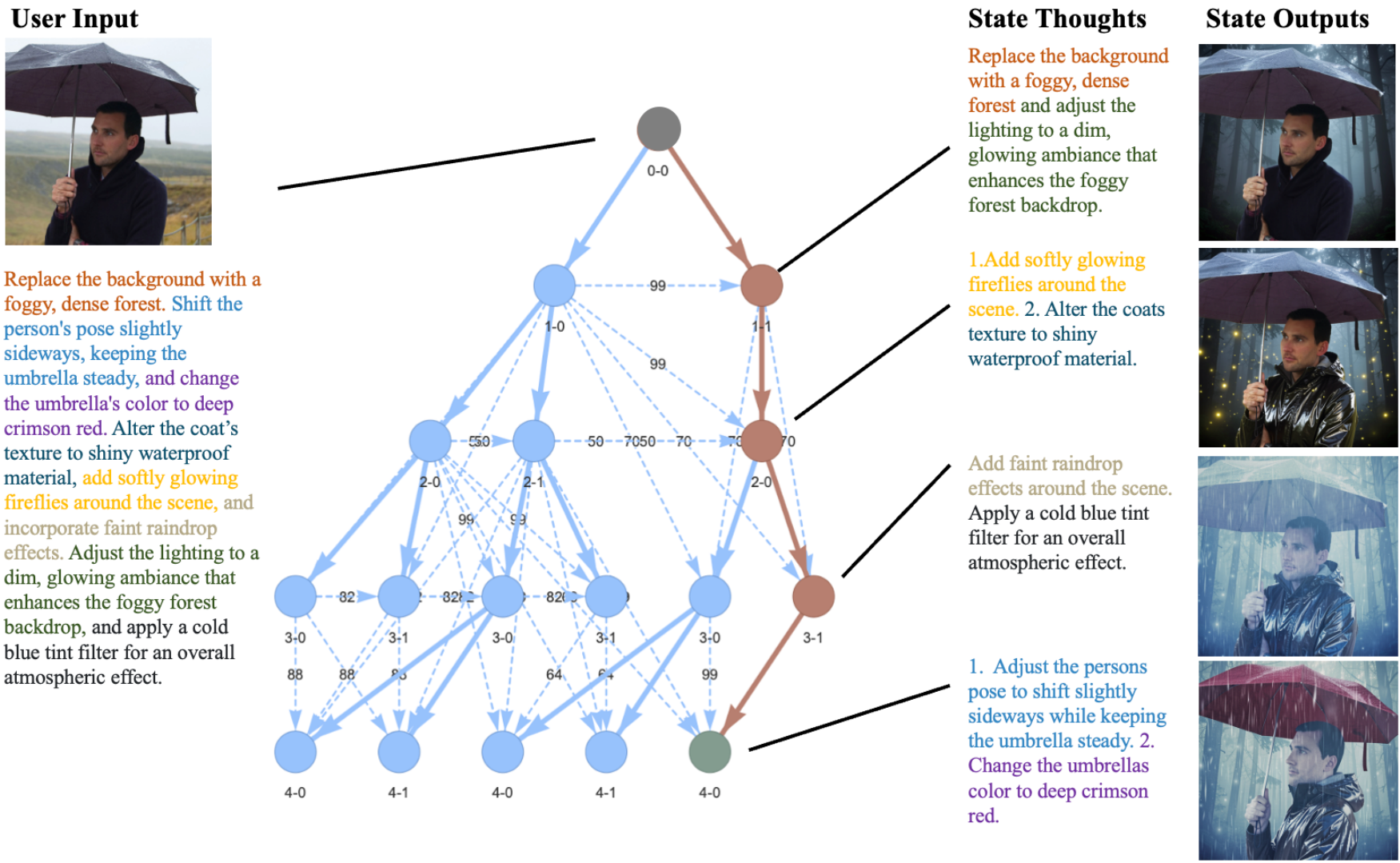}

\includegraphics[width=11cm]{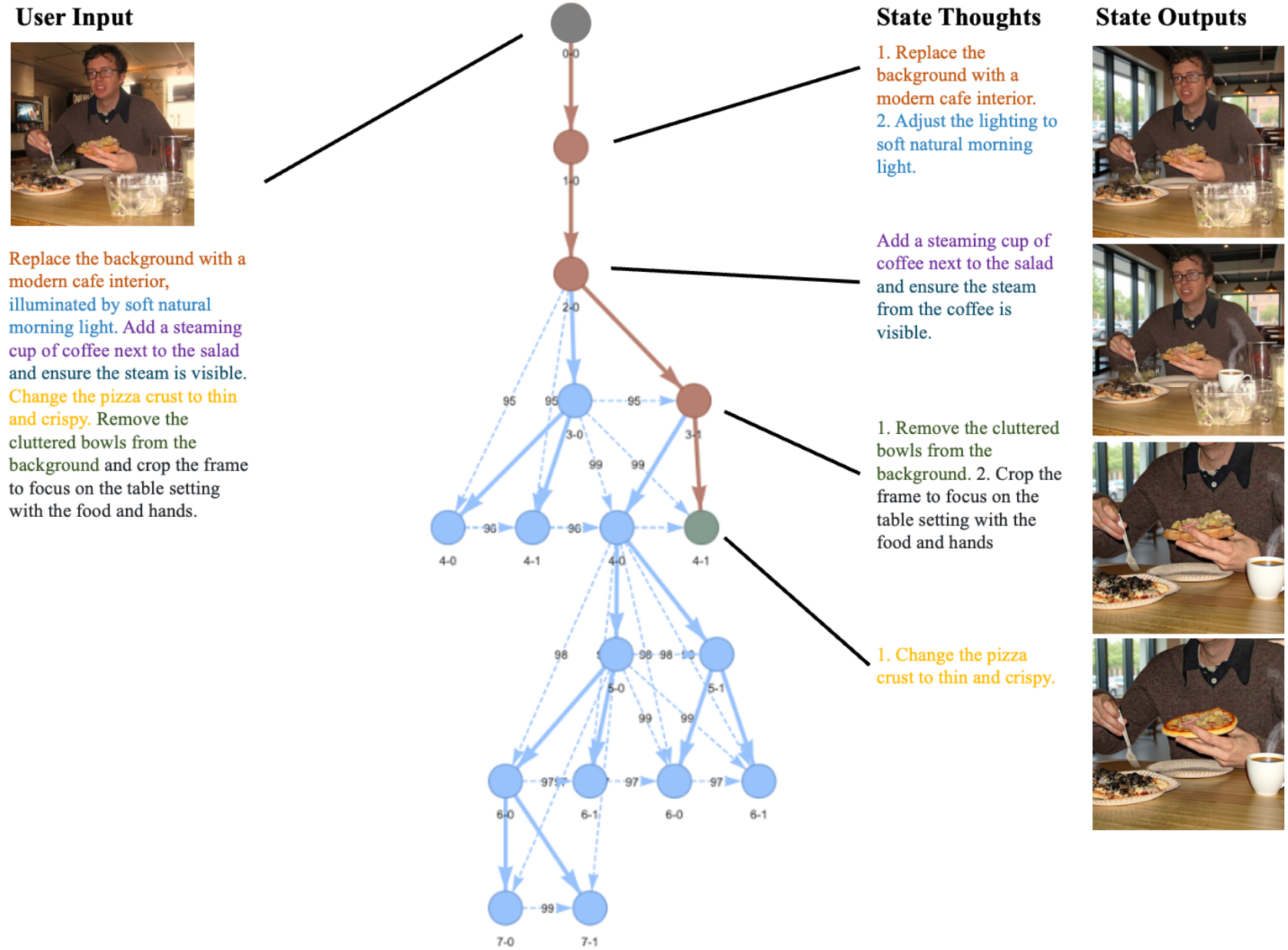}
  \caption{Visualizable Inference Topology. Task decomposition constructs diverse decision paths that lead to different editing results. Visualization of the inference topology show this process in a transparent way, enabling information check of every state.
  }
  \label{fig:InferenceTopology}
\end{figure}

We provide more examples of MSRAMIE structured multimodal reasoning process in \Cref{fig:InferenceTopology}. For each case, it enables transparent decision trajectories leading to the final editing results. All the intermediate results and decision rationale are accessible for the user to enhance user friendliness and controllability. It is worth mentioning that exploration on different branches is important as deep inference is not a must for the best results.

\subsection{Multi-instruction Degradation}
\begin{figure}[!h]
\centering
\includegraphics[width=12.5cm]{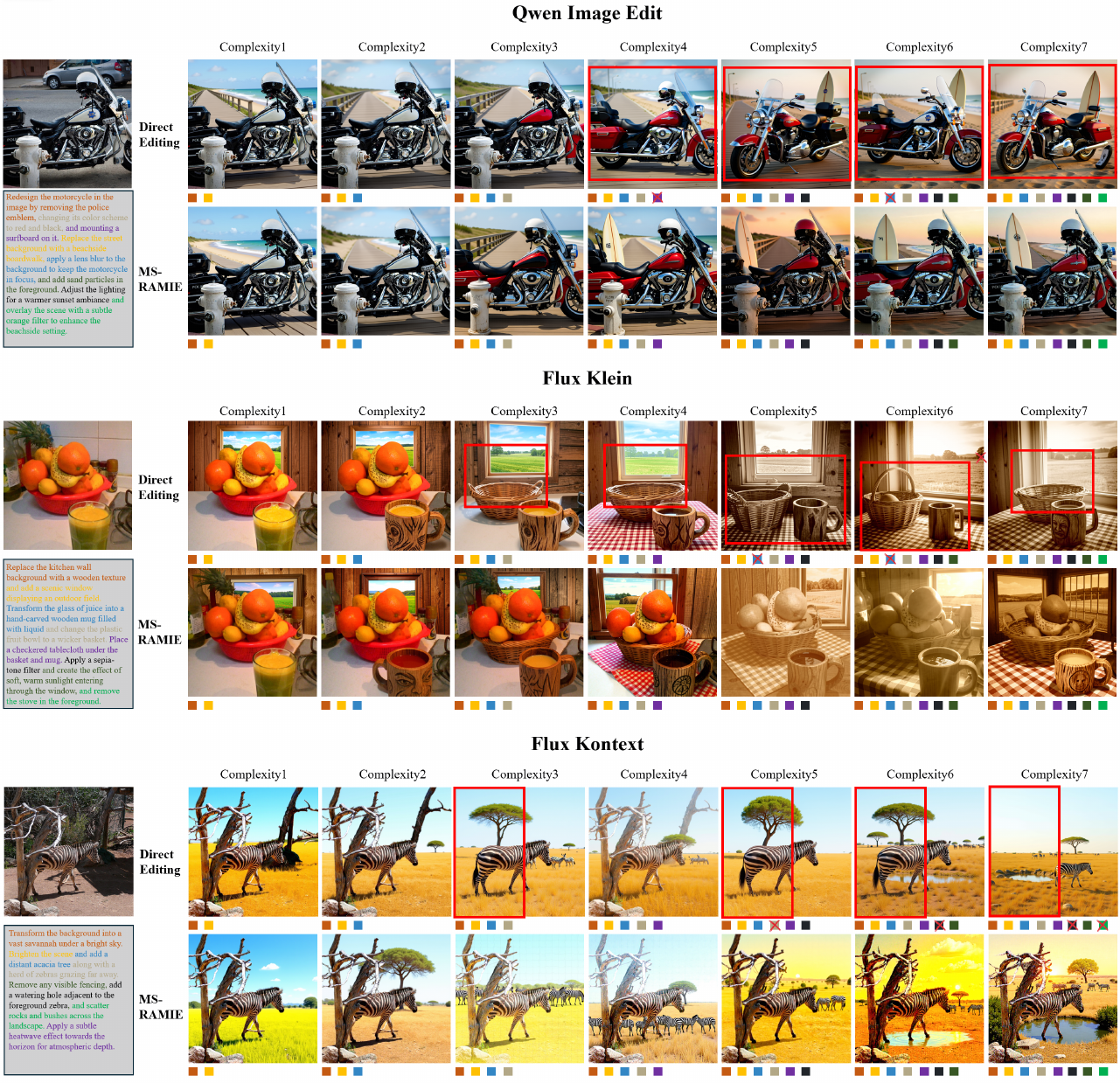}
  \caption{Performance Degradation Across Complexities \& Enhancement by MSRAMIE. Colored blocks beneath each image denote individual instruction items, and a red cross indicates failure to satisfy the corresponding item. Regions where visual content of original images is not preserved in Direct Editing but retained by MSRAMIE are highlighted with red boxes.
  }
  \label{fig:Degradation}
\end{figure}

As in \Cref{fig:Degradation}, we present visual comparisons to re-emphasize the consistent performance degradation of existing instruction-based image editing models in multi-instruction scenarios, manifested as missed editing requirements (poor Instruction Following) and unintended modifications (poor Identity Preservation).

The underlying cause is intuitive. Complex multi-instructions typically involve multiple visual elements across different regions of the image. Consequently, (1) the model must attend to large portions of the image, during which many elements that should remain unchanged are inadvertently and incorrectly involved in the editing area, resulting in unwanted modification; and (2)multiple target elements compete for attention during the cross-modal attention conditioning, causing some to receive insufficient attention scores and be ignored, resulting in missed requirements.

MSRAMIE mitigates this issue through its inherent divide-and-conquer strategy based on task decomposition. By decomposing a multi-instruction editing task into a sequence of simpler editing steps, each step involves fewer visual elements and focuses on a more localized editing region. This improves both instruction following and identity preservation.

\clearpage
\section*{Appendix G Discussions}
\label{sec:AppendixG}
\renewcommand{\thesubsection}{\arabic{subsection}}
\setcounter{subsection}{0}
\setcounter{figure}{0}
\setcounter{table}{0}
\subsection{Regarding Generalization}
In this study, we selected three representative instruction-based image editing models, Qwen-Image-Edit~\cite{wu2025qwenimage}, Flux-Kontext~\cite{bfl2025fluxkontext}, and Flux2-Klein~\cite{flux2techreport}, for the following reasons.

First, earlier models such as InstructPix2Pix~\cite{brooks2023instructpix2pix} and MagicBrush~\cite{zhang2023magicbrush} exhibit limited robustness even for single-instruction editing, which constitutes the atomic unit of the multi-instruction setting considered in this work. Our method targets models that already perform well on single-instruction editing but degrade when handling multiple instructions; therefore, those earlier models fall outside the scope of our evaluation.

Second, we aimed to cover the major architectural paradigms in instruction-based image editing. The selected models represent the two mainstream architectural designs: (1) Qwen-Image-Edit~\cite{wu2025qwenimage} as a diffusion-based multimodal diffusion-transformer (MMDiT)\cite{peebles2023scalable} architecture with dual image encoding; (2) Flux-Kontext~\cite{bfl2025fluxkontext} and Flux2-Klein~\cite{flux2techreport} as unified rectified-flow transformer\cite{lipman2022flow} that performs generation and editing through in-context conditioning with sequence concatenation of textual and visual inputs. Evaluating MSRAMIE on these diverse architectures demonstrates its strong generalization capability across mainstream  modern instruction-based image editing frameworks.

\subsection{Regarding Framework and Modules Design}
The design of the Evaluator submodule within the \textit{Instructor} top-level module is particularly critical. A reliable and scaled evaluation signal is essential for efficient topology search and reliable hyperparameter tuning. In early experiments, we attempted MLLM-based scoring, where the model assigns a numerical score (e.g., 0–10) to each state. However, this approach proved unstable because such scores lack a consistent reference scale. In contrast, VQAScore~\cite{lin2024vqascore} provides a more reliable signal due to its bounded 0–1 range and its interpretation as an “affirmative answer proportion”, which enables more stable search process.

Regarding search strategy, depth-first search (DFS) is more budget-efficient than breadth-first search (BFS). As instruction following (IF) is typically prioritized over other metrics in instruction-based image editing, due to task decomposition, it requires large inference depth to satisfy all the editing requirements. DFS therefore enables faster discovery of feasible solutions under limited inference budgets. Nevertheless, exploring more advanced search algorithms, such as A* or beam search, remains an interesting direction for future work.

\subsection{Regarding Explicit Topology Search}
Explicit topology-structured search is sometimes considered inefficient and unacceptable for tasks requiring real-time responses. However, in instruction-based image editing, real-time responding is usually not a must, and such an explicit search can improve user experience by providing a transparent decision process. This transparency allows users to inspect intermediate states and potentially select or reuse partial solutions.

Furthermore, explicit search enables real-time monitoring of the reasoning trajectory, making it possible to terminate undesirable exploration paths early or manually guide the search toward more promising directions.

\clearpage
\par\vfill\par
\end{document}